\documentclass[11pt]{article}

\usepackage[preprint]{acl}
\usepackage{times}
\usepackage{latexsym}
\usepackage[T1]{fontenc}
\usepackage[utf8]{inputenc}
\usepackage{microtype}
\usepackage{inconsolata}
\usepackage{graphicx}
\usepackage{booktabs}
\usepackage{multirow}
\usepackage{adjustbox}
\usepackage{xcolor}
\usepackage[section]{placeins}
\usepackage{pifont} % for xmark
\usepackage{threeparttable}
\usepackage{float}
\usepackage{newunicodechar}
\usepackage{fontspec}
\newunicodechar{ẹ}{\d{e}}\newunicodechar{ọ}{\d{o}}
\newunicodechar{ụ}{\d{u}}\newunicodechar{ị}{\d{i}}

\usepackage{fontspec}
\newfontfamily\tifinaghfont{NotoSansTifinagh-Regular.ttf}[Scale=MatchUppercase]

\newcommand{\cmark}{\ding{51}}%
\newcommand{\xmark}{\ding{55}}%

\definecolor{posgreen}{rgb}{0.0, 0.50, 0.0}
\definecolor{negred}{rgb}{0.70, 0.0, 0.0}
\newcommand{\pv}[1]{{\footnotesize\textcolor{posgreen}{+#1}}}
\newcommand{\nv}[1]{{\footnotesize\textcolor{negred}{--#1}}}

\title{An Empirical Study of Many-Shot In-Context Learning for Machine Translation of Low-Resource Languages}

\author{
Yinhan Lu$^{1,2}$\thanks{These authors contributed equally.}, Gaganpreet Jhajj$^{1,3}$\footnotemark[1], Chen Zhang$^{4}$, Anietie Andy$^{5}$, David Ifeoluwa Adelani$^{1,2,5}$ \\
$^{1}$Mila -- Quebec AI Institute \quad $^{2}$McGill University \quad $^{3}$Athabasca University \\
$^{4}$Peking University \quad $^{5}$Howard University \quad $^{5}$Canada CIFAR AI Chair \\
\texttt{\{yinhan.lu, gaganpreet.jhajj, david.adelani\}@mila.quebec} \\
\texttt{zhangch@pku.edu.cn} \quad \texttt{anietie.andy@howard.edu}
}

\begin{document}
\maketitle

% ================================================================
% ABSTRACT
% ================================================================
\begin{abstract}
In-context learning (ICL) allows large language models (LLMs) to adapt to new tasks from a few examples, making it promising for languages underrepresented in pre-training. Recent work on many-shot ICL suggests that modern LLMs can further benefit from larger ICL examples enabled by their long context windows. However, such gains depend on careful example selection, and the inference cost can be prohibitive for low-resource language communities. In this paper, we present an empirical study of many-shot ICL for machine translation from English into ten truly low-resource languages recently added to FLORES+. We analyze the effects of retrieving more informative examples, using out-of-domain data, and ordering examples by length. Our findings show that many-shot ICL becomes more effective as the number of examples increases. More importantly, we show that BM25-based retrieval substantially improves data efficiency: 50 retrieved examples roughly match 250 many-shot examples, while 250 retrieved examples perform similarly to 1,000 many-shot examples. We further show that ICL provides additional gains on top of fine-tuning.
\end{abstract}

% ================================================================
% §1 INTRODUCTION
% ================================================================
\section{Introduction}
\label{sec:intro}
% David re-writing introduction
Large Language Models (LLMs) have shown strong performance on many natural language processing (NLP) tasks, including generation tasks such as summarization and machine translation (MT) in multilingual settings~\cite{pu2023summarization,zhu-etal-2024-multilingual,team2024gemini}. However, their performance remains limited for very low-resource and endangered languages.

In-context learning (ICL)~\cite{brown2020language,dong2024survey}, which leverages few-shot examples, is one of the most promising emergent abilities of LLMs and has shown strong potential for low-resource languages (LRLs)~\cite{lin-etal-2022-shot,zhang2023prompting}. However, performance does not necessarily improve as the number of few-shot examples increases, leading many studies to limit the number of shots (e.g., to less than 20)~\cite{zhang_teaching_2024,tanzer2024a,court-elsner-2024-shortcomings}. This issue has led to various streams of work focusing on choosing the best few-shot examples, either through similarity between examples, retrieving the most similar to target examples, domain effects, and use of pseudo parallel data~\citep{agrawal-etal-2023-context,marashian2025,pei_2025_understanding}. However, most of these studies do not cover LRLs where retrieval is often weaker.

Recently, a new paradigm has emerged that focuses on prompting with large numbers of examples, or ``many-shot'' prompting (e.g., 1,000 examples)~\cite{agarwal_many-shot_2024}. This approach seems beneficial for LRLs that had limited representation during pretraining, and can make better use of a few thousand examples than fine-tuning, which typically requires more data~\cite{adelani-etal-2022-thousand,vieira-etal-2024-much}. However, despite its promise, the high inference cost makes it challenging for practical use. Exploring more effective many-shot example selection could improve its scalability for low-resource language communities.

% --- Figure 1: Scaling Curves (combined) ---
\begin{figure*}[t]
\centering
\includegraphics[width=\textwidth]{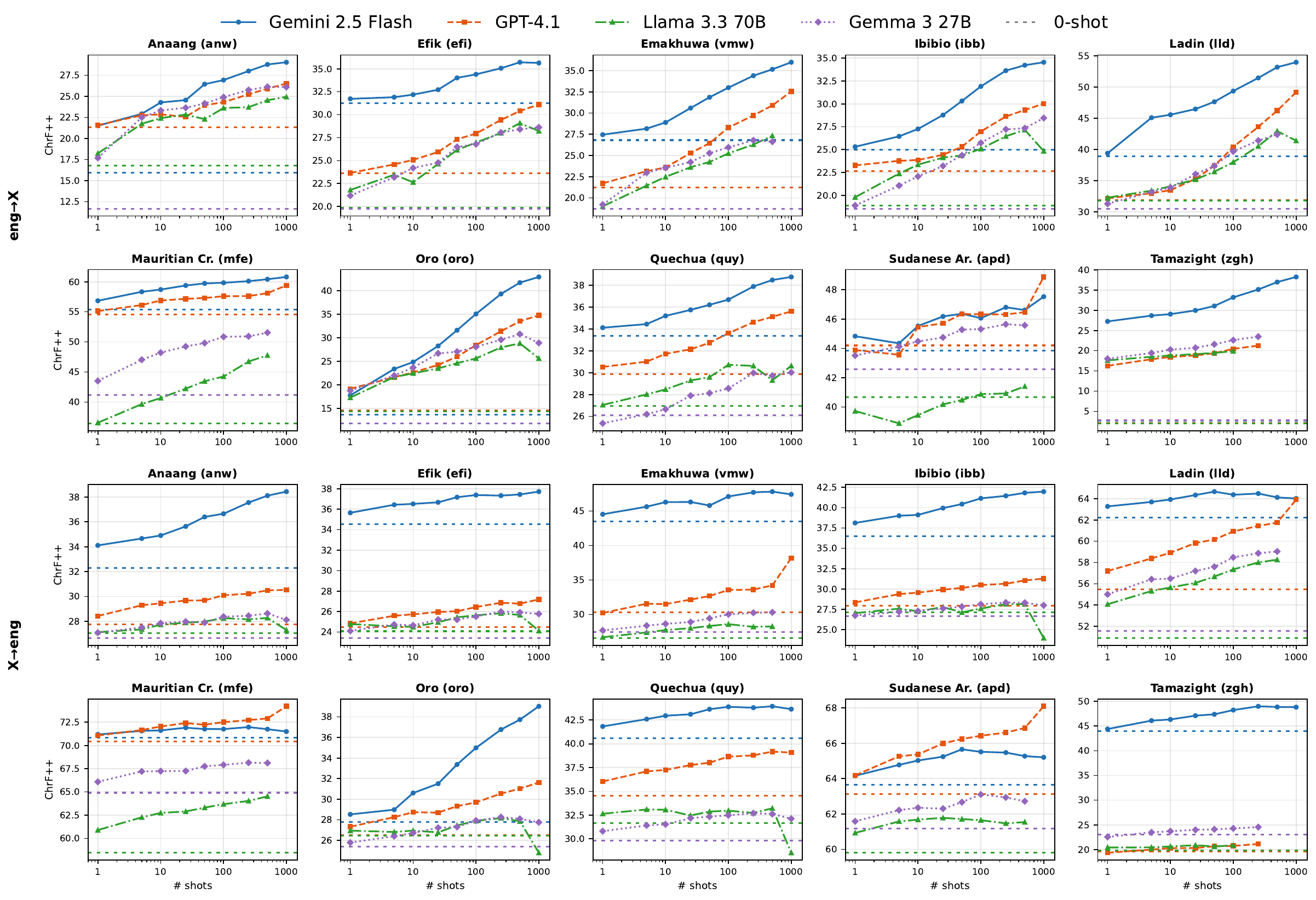}
\vspace{-5mm}
\caption{Per-language scaling curves (chrF++, random selection). Top two rows: eng$\rightarrow$X; bottom two rows: X$\rightarrow$eng. }
\label{fig:scaling_chrf}
\end{figure*}
% The corresponding spBLEU results are shown in \autoref{fig:scaling}.

In this paper, we investigate the effectiveness of many-shot machine translation (MT) from English into 10 truly LRLs %and endangered languages 
with limited exposure during LLM pretraining. We select these languages carefully based on the recency of the benchmarks, focusing on newly added languages in FLORES+. 
We further examine how to reduce inference costs by retrieving a smaller set of more effective many-shot examples using simple methods such as BM25. Besides, we study the impact of out-of-domain examples, such as religious texts, on in-domain translation performance for Wikipedia, as well as the effect of example ordering from short to long, inspired by curriculum learning. We finally compare many-shot ICL against fine-tuning of an open-weight model (\texttt{Gemma~3~27B}) .

Our evaluation on both open-weight and proprietary models, such as \texttt{Gemini 2.5 Flash} and \texttt{GPT-4.1}, shows the effectiveness of the many-shot approach on these languages, where performance more than doubles in some languages with more examples. Similarly, leveraging a simple BM25 retrieval from English examples drastically reduces cost, where BM25 with 50 examples roughly matches 250 many-shot examples, and BM25 with 250 examples has similar performance as 1,000 many-shot examples. We find that many shots from within the same domain are consistently better for most languages; however, some languages are not affected by this domain mismatch, which shows some promise in the case of absence of in-domain examples. Furthermore, the ordering of examples does not seem to have a strong effect on many-shot performance in our experiments. Finally, we show that ICL with only 10--25 retrieved examples  on open-source models without fine-tuning matches fine-tuning on the full 1{,}012-sentence pool, and increasing the number of samples continues to improve performance even on top of fine-tuning. 
% ================================================================
% §3 EXPERIMENTAL SETUP
% ================================================================
\section{Experimental Setup}
\label{sec:setup}

% --------------------------------------------------
% \subsection{Languages}
% \label{sec:languages}

\vspace{1mm}
\noindent \textbf{Focus Languages} 
 
We focus on ten extremely low-resource languages that were recently added to FLORES+ %\footnote{\url{https://huggingface.co/datasets/openlanguagedata/flores_plus}} 
or extended from FLORES-200~\citep{nllb-24}, including four Nigerian languages (Anaang, Efik, Ibibio, and Oro), Sudanese Arabic, Emakhuwa, Ladin, Mauritian Creole, Tamazight, and Quechua\footnote{While Tamazight and Quechua were originally part of FLORES-200, Tamazight is improved by the community.}. 
Eight languages use the Latin script; the exceptions are Tamazight
(Tifinagh) and Sudanese Arabic (Arabic script).
\autoref{app:languages} provides more details.
\begin{table*}[t]
\centering
\small
\resizebox{\textwidth}{!}{%
\begin{tabular}{cc|cccccccccc|c}
\toprule
%\textbf{Shots} & \textbf{Method} & anw & apd & efi & ibb & lld & mfe & oro & quy & vmw & zgh & \textbf{Avg} \\
\textbf{Shots} & \textbf{Method} & Anaang & Sudanese & Efik & Ibibio & Ladin & Mauritian & Oro & Quechua & Emakhuwa & Tamazight &  \textbf{Avg} \\

\midrule
\multicolumn{13}{c}{\textit{Gemini 2.5 Flash, eng $\rightarrow$ X (chrF++)}} \\
\midrule
\multirow{3}{*}{5}    & BM25   & \textbf{25.8} & \textbf{45.7} & \textbf{33.5} & \textbf{28.7} & \textbf{47.6} & \textbf{58.6} & \textbf{28.6} & \textbf{35.9} & \textbf{30.9} & \textbf{32.0} & 36.7 \\
                       & Random & 22.9 & 44.3 & 31.9 & 26.4 & 45.1 & 58.3 & 23.3 & 34.5 & 28.2 & 28.7 & 34.4 \\
                       & $\Delta$ & \pv{2.9} & \pv{1.4} & \pv{1.6} & \pv{2.3} & \pv{2.5} & \pv{0.3} & \pv{5.3} & \pv{1.4} & \pv{2.7} & \pv{3.3} & \pv{2.4} \\
\midrule
\multirow{3}{*}{50}   & BM25   & \textbf{28.2} & \textbf{46.9} & \textbf{35.2} & \textbf{32.8} & \textbf{51.8} & \textbf{60.3} & \textbf{38.8} & \textbf{38.0} & \textbf{34.2} & \textbf{36.4} & 40.3 \\
                       & Random & 26.4 & 46.4 & 34.0 & 30.3 & 47.6 & 59.7 & 31.6 & 36.2 & 31.9 & 31.1 & 37.5 \\
                       & $\Delta$ & \pv{1.8} & \pv{0.5} & \pv{1.2} & \pv{2.5} & \pv{4.2} & \pv{0.6} & \pv{7.2} & \pv{1.8} & \pv{2.3} & \pv{5.3} & \pv{2.7} \\
\midrule
\multirow{3}{*}{250}  & BM25   & \textbf{28.9} & \textbf{46.8} & \textbf{35.4} & \textbf{34.2} & \textbf{53.2} & \textbf{60.5} & \textbf{41.8} & \textbf{38.6} & \textbf{35.3} & \textbf{37.7} & 41.2 \\
                       & Random & 28.0 & \textbf{46.8} & 35.1 & 33.6 & 51.4 & 60.1 & 39.3 & 37.9 & 34.4 & 35.1 & 40.2 \\
                       & $\Delta$ & \pv{0.9} & 0.0 & \pv{0.3} & \pv{0.6} & \pv{1.8} & \pv{0.4} & \pv{2.5} & \pv{0.7} & \pv{0.9} & \pv{2.6} & \pv{1.1} \\
\midrule
\multirow{3}{*}{1,000} & BM25   & \textbf{29.5} & 47.0 & \textbf{35.9} & 34.4 & \textbf{54.1} & \textbf{61.1} & \textbf{43.2} & \textbf{38.8} & 35.7 & \textbf{38.3} & 41.8 \\
                       & Random & 29.0 & \textbf{47.5} & 35.6 & \textbf{34.5} & 53.9 & 60.8 & 42.9 & \textbf{38.8} & \textbf{36.0} & 38.2 & 41.7 \\
                       & $\Delta$ & \pv{0.5} & \nv{0.5} & \pv{0.3} & \nv{0.1} & \pv{0.2} & \pv{0.3} & \pv{0.3} & 0.0 & \nv{0.3} & \pv{0.1} & \pv{0.1} \\
\bottomrule
\end{tabular}
}
\vspace{-2mm}
\caption{BM25 vs.\ Rand (chrF++, eng$\rightarrow$X, Gemini). \textbf{Bold}: winner. $\Delta$: \textcolor{posgreen}{green} = BM25 better, \textcolor{negred}{red} = Random better.}
\label{tab:bm25_vs_random}
\end{table*}

\vspace{1mm}
\noindent \textbf{Experimental Design} 
% \paragraph{Experimental Design}
% \subsection{Experimental Design} 
For each test sentence, we construct a prompt containing a task instruction, $k$ parallel example pairs (source and target), and the test sentence as a query; the model then generates the translation in one pass.
We test $k \in \{0, 1,5, 10, 25, 50, 100, 250, 500, 1000\}$; the full prompt template appears in Appendix~\ref{app:prompts}. Our evaluation targets three research questions. %. % posed in \S\ref{sec:intro}. 

% \vspace{1mm}
% \noindent \textbf{(1) Scaling} 
% \paragraph{(1) Scaling} 
\textbf{(1) Scaling} 
We evaluate with randomly sampled examples across several $k$, and ten languages in two directions (``eng$\rightarrow$X'' and ``X$\rightarrow$eng''), using \texttt{Gemini 2.5 Flash} \cite{gemini25}, \texttt{GPT-4.1},  \texttt{Llama 3.3 70B}~\citep{grattafiori2024llama} and \texttt{Gemma 3 27B}~\citep{gemmateam2025gemma3technicalreport}. 

% \vspace{1mm}
% \noindent \textbf{(2) Example Selection} 
% \paragraph{(2) Example Selection} 
\textbf{(2) Example Selection} 
We compare selecting ICL examples randomly vs. BM25~\cite{10.1561/1500000019} retrieval on the English source side and therefore restrict this experiment to eng$\rightarrow$X. We report other  retrieval models such as Qwen Embedding model in~\autoref{app:ordering}.

% \vspace{1mm}
% \noindent \textbf{(3) Domain Mismatch} 
% \paragraph{(3) Domain Mismatch} 
\textbf{(3) Domain Mismatch} 
On the 7 languages where Bible translations exist\footnote{We excluded Sudanese Arabic Bible following native speaker feedback that it is too similar to Standard Arabic.} (Table~\ref{tab:languages}), we compare Bible examples against in-domain BM25 and random selection, all in the ``eng$\rightarrow$X'' direction. We retrieve Bible examples with BM25, the same setup as the in-domain experiments, so differences in performance reflect domain rather than retrieval. This tests whether out-of-domain data benefits from scaling the way in-domain data does.
% ================================================================
% §4 RESULTS

\section{Results}
\label{sec:results}
Here we mainly report results with chrF++; results with spBLEU, shown in \autoref{app:full_results}, lead to similar conclusions.
% ~\citep{popovic-2017-chrf}

% --------------------------------------------------
% §4.1 SCALING
% --------------------------------------------------
\subsection{Effect of Scaling Many-Shot Examples}
\label{sec:scaling}

\autoref{fig:scaling_chrf} shows the results for different numbers of ICL examples from $k=1,5,..,1000$. The results show consistent improvement across all LLMs, especially the proprietary models (\texttt{Gemini 2.5 Flash} and \texttt{GPT-4.1}). In the ``eng$\rightarrow$X'' direction, \texttt{Gemini 2.5 Flash} achieves the best overall results across all shot settings, with performance gains ranging from $3.7$ to $35.6$ from 0-shot to 1,000-shot. Performance is also moderately higher when comparing a few-shot setting, say 10-shot to a many-shot setting, say 1,000-shot, with gains of $2.0$ to $18.1$. 

Tamazight benefited the most from few-shot (1-shot) while Oro benefited the most from many-shot; We discuss possible explanations and provide human evaluation in \autoref{app:qualitative}. In contrast, the ``X$\rightarrow$eng'' direction shows significantly less improvement than generation into a LRL. However, the trend remains consistently positive except for Sudanese Ar. and Mauritian Cr., which may reflect confusion with related dominant languages such as Arabic and French.

\vspace{1mm}
\noindent \textbf{Open weight models experience in-consistencies in higher many-shot.} 

\texttt{Gemma 3 27B} and \texttt{Llama 3.3 70B} sometimes fail at $k=500$ or $k=1,000$ due to limited context window, and weaker ability to process large numbers of ICL examples. Many-shot is still most effective for proprietary models, aligning with the base LLM's competence.

\vspace{1mm}
\noindent \textbf{Larger many-shot does not always lead to the best performance.} 

In some cases, $k=250$ or $k=500$ leads to better results than $k=1,000$, especially in the ``X$\rightarrow$eng'' direction, while for some other languages the performance keeps increasing.

\begin{figure*}[t]
\centering
\includegraphics[width=0.98\textwidth]{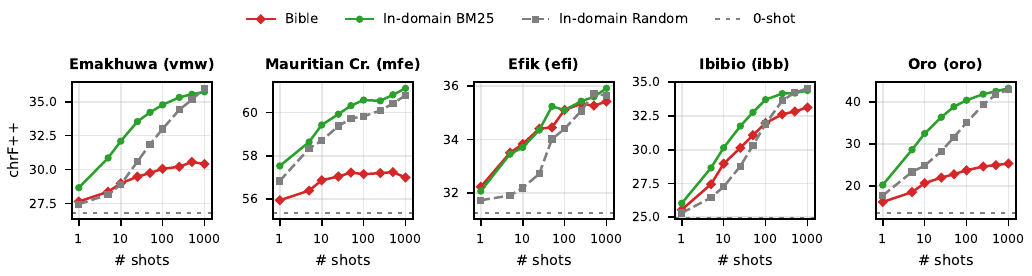}
\vspace{-4mm}
\caption{Bible vs.\ in-domain examples (chrF++, eng$\rightarrow$X, Gemini 2.5 Flash). Bible examples plateau or degrade with more shots, while in-domain examples scale consistently.}
\label{fig:bible_vs_indomain}
\vspace{-3mm}
\end{figure*}

\subsection{Discussion}
\vspace{1mm}
\noindent \textbf{Example Selection: BM25 vs.\ Random}
% \paragraph{Example Selection: BM25 vs.\ Random}
\label{sec:selection}
\autoref{tab:bm25_vs_random} compares random sampling of ICL examples to BM25 retrieval on the English source side (eng$\rightarrow$X).\footnote{Applied to the English side because applying BM25 to LRLs may yield inconsistent retrieval across languages.}  BM25 outperforms random selection for nearly all languages, especially at low shot counts, although the difference narrows as more examples are added. Most importantly, it is more data efficient: 50-shot with BM25 achieves the same average performance as 250-shot of Random ICL ($\approx 40.2$), and 250-shot BM25 matches 1,000-shot random ICL. This provides more practical usability of many-shot for low-resource communities, drastically reducing inference cost.  

\vspace{1mm}
\noindent \textbf{Effect of Out-of-Domain Examples}
% \paragraph{Effect of Out-of-Domain Examples} 
\autoref{fig:bible_vs_indomain} shows that many-shot ICL from within the same domain (Wikipedia) is more effective than that of out-of-domain (Bible) across all languages. However, some languages do benefit more from the religious domain than others, such as Efik and Ibibio. This finding suggests that all hope is not lost if in-domain examples are not available; out-of-domain examples may still be useful when they remain distributionally close to in-domain text—a closeness that depends on the language property.

\begin{figure}[t]
\centering
\includegraphics[width=0.47\textwidth]{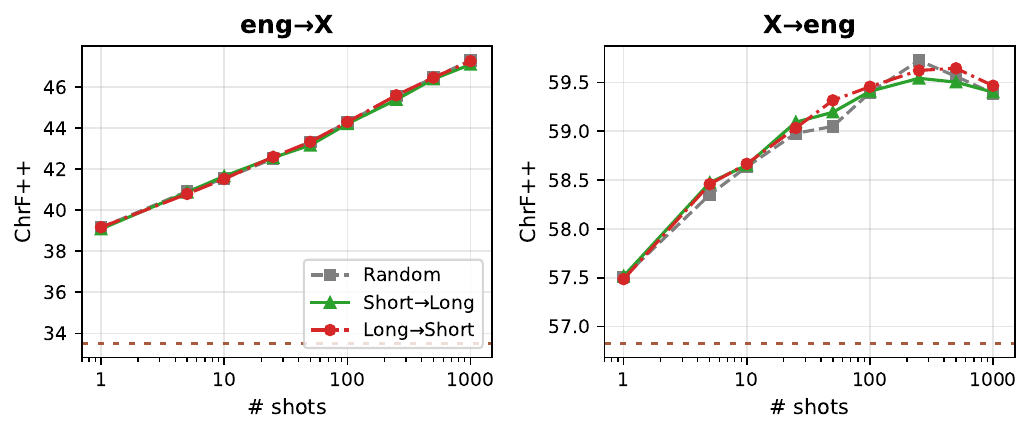}
\vspace{-3mm}
\caption{Effect of example ordering on MT (chrF++, Gemini~2.5~Flash, averaged across vmw, zgh, lld, mfe, and apd). \textbf{S2L} and \textbf{L2S} sort examples by source length.}
\label{fig:ordering_length}
\end{figure}

\vspace{1mm}
\noindent \textbf{Effect of Ordering ICL Examples} \quad
% \paragraph{Effect of Ordering ICL Examples} 
\autoref{fig:ordering_length} shows the results of ordering examples from Short-to-Long (S2L) and Long-to-Short (L2S). We did not observe any significant impact of ordering by length, especially in  eng$\rightarrow$X. Semantic relevance via BM25 retrieval seems to be more important.

\begin{table}[t]
\centering
\small
\setlength{\tabcolsep}{6pt}
\begin{tabular}{l|cccccc}
\toprule
\textbf{Variant} & \textbf{0} & \textbf{5} & \textbf{10} & \textbf{25} & \textbf{50} & \textbf{250} \\
\midrule
Base \texttt{Gemma}    & 19.8 & 25.5 & 27.0 & 28.4 & 29.2 & 29.3 \\
+ \textsc{\textsc{LoRA}$_{250}$}   & 24.1 & 26.3 & 27.0 & 28.2 & 28.9 & 29.4 \\
+ \textsc{\textsc{LoRA}$_{1012}$}  & 27.3 & 28.5 & 28.9 & 29.3 & 29.7 & 30.1 \\
\bottomrule
\end{tabular}
\caption{LoRA fine-tuning combined with many-shot ICL on \mbox{eng$\rightarrow$Efik} (chrF++): columns are BM25-retrieved in-context shots.}
\label{tab:lora_budget_main}
\end{table}

\vspace{1mm}
\noindent \textbf{Comparison with Fine-tuning} \quad
% \paragraph{Comparison with Fine-tuning}
\label{sec:lora}
Fine-tuning is an alternative to many-shot ICL for open-weight users. We LoRA fine-tune \texttt{Gemma~3~27B} on the full 1{,}012-sentence pool (\textsc{LoRA}$_{1012}$) and on a 250-sentence BM25-selected subset (\textsc{LoRA}$_{250}$) for \mbox{eng$\rightarrow$\{Efik, Ibibio\}}, evaluated at 0 to 250 BM25 shots.
% (App.~\ref{app:lora}). 
% We discuss \mbox{eng$\rightarrow$Efik} for concreteness; \mbox{eng$\rightarrow$Ibibio} follows the same qualitative pattern.
%\textbf{LoRA alone is comparable to a modest retrieved prompt.} 
For \mbox{eng$\rightarrow$Efik}, without ICL examples, \textsc{LoRA}$_{1012}$ reaches 27.3 chrF++, which is slightly below base \texttt{Gemma~3~27B} with 25 shots (28.4), while \textsc{LoRA}$_{250}$ reaches 24.1, slightly below the base model with 5 shots (25.5). LoRA and ICL are complementary, but the marginal benefit of LoRA quickly narrows. Adding ICL on top of LoRA still improves performance, even though the examples are drawn from the same pool used for fine-tuning. 
%; however, by 25 shots the base model already matches \textsc{LoRA}$_{250}$ (28.4 vs.\ 28.2) and trails \textsc{LoRA}$_{1012}$ by only 0.9 chrF++ points (28.4 vs.\ 29.3). With more shots, the gap between base and the fine-tuned models largely disappears, 
While fine-tuning is helpful before ICL, open-weight models still lag behind the closed-source baseline: \texttt{Gemini~2.5~Flash} with only 5 shots already exceeds \textsc{LoRA}$_{1012}$ at 250 shots (33.5 vs.\ 30.1). Also, given the huge parameter sizes of proprietary LLMs, fine-tuning is typically not an option, but ICL can be always be leveraged. \mbox{eng$\rightarrow$Ibibio} follows the same qualitative pattern. See full result of fine-tuning in \autoref{app:lora}.

% ================================================================
% §6 CONCLUSION
% ================================================================
\section{Conclusion}
In this paper, we study many-shot in-context learning for MT across ten LRLs, scaling from 0 to 1,000 parallel examples and evaluating four LLMs. 
Performance improves roughly log-linearly as the number of ICL examples increases, with larger gains for translation into LRLs (eng$\rightarrow$X) than for X$\rightarrow$eng, which shows only moderate improvements.
For example selection, retrieval based on similarity on the English side improves sample efficiency: 50 BM25-retrieved examples match the performance of 250 randomly selected ones, and 250 BM25 examples match 1,000 random ICL examples.
Out-of-domain ICL examples yield smaller overall gains than in-domain examples, although they remain beneficial for some LRLs.

% ================================================================
% LIMITATIONS
% ================================================================
\section*{Limitations}

All language pairs in this study involve English as either the source or target language; we did not evaluate non-English-centric translation, which may exhibit different scaling behavior.

Furthermore, we used a single fixed prompt template and did not explore prompt engineering or chain-of-thought strategies, which could affect scaling in ways we have not measured. Additionally, our evaluation is limited to automatic metrics (spBLEU, chrF++); human evaluation would provide a more complete picture of translation quality, particularly for languages where automatic metrics may be less reliable. We also did not explore embedding-based metrics such as COMET~\citep{rei-etal-2020-comet} and MetricX~\citep{juraska-etal-2023-metricx} because the languages studied are extremely low-resource and not covered by these metrics.

More broadly, parallel data for extremely low-resource languages may contain annotation errors or inconsistencies, which can affect both the quality of in-context examples and the reliability of reference translations for these automatic evaluation. Hence, translation outputs for these languages should be verified by native speakers before deployment.

Finally, it is important to note that the full set of experiments in this paper required more than \$30,000 in API credits, highlighting a significant barrier to reproducibility for low-resource language communities. To mitigate this, we release all results and analyses to reduce the need for others to replicate these costly runs.

\section*{Acknowledgment}
This research was supported IVADO and the Canada First Research Excellence Fund. We are grateful for the support of the Azure sponsorship credits granted by Microsoft’s AI for Good Research Lab, which enabled us to carry out computationally expensive inference.

%================================================================
% BIBLIOGRAPHY
% ================================================================
%\clearpage
% \nocite{*} % TODO: remove once real \cite commands are added
\bibliography{custom}

@inproceedings{
samil2026sudaneseflores,
title={Sudanese-Flores: Extending {FLORES}+ to Sudanese Arabic Dialect},
author={Hadia Mohmmedosman Ahmed Samil and David Ifeoluwa Adelani},
booktitle={7th Workshop on African Natural Language Processing},
year={2026},
url={https://openreview.net/forum?id=uLKTcetdkB}
}

@article{10.1561/1500000019,
author = {Robertson, Stephen and Zaragoza, Hugo},
title = {The Probabilistic Relevance Framework: BM25 and Beyond},
year = {2009},
issue_date = {April 2009},
publisher = {Now Publishers Inc.},
address = {Hanover, MA, USA},
volume = {3},
number = {4},
issn = {1554-0669},
url = {https://doi.org/10.1561/1500000019},
doi = {10.1561/1500000019},
journal = {Found. Trends Inf. Retr.},
month = apr,
pages = {333–389},
numpages = {57}
}

@inproceedings{dong2024survey,
    title = "A Survey on In-context Learning",
    author = "Dong, Qingxiu  and
      Li, Lei  and
      Dai, Damai  and
      Zheng, Ce  and
      Ma, Jingyuan  and
      Li, Rui  and
      Xia, Heming  and
      Xu, Jingjing  and
      Wu, Zhiyong  and
      Chang, Baobao  and
      Sun, Xu  and
      Li, Lei  and
      Sui, Zhifang",
    editor = "Al-Onaizan, Yaser  and
      Bansal, Mohit  and
      Chen, Yun-Nung",
    booktitle = "Proceedings of the 2024 Conference on Empirical Methods in Natural Language Processing",
    month = nov,
    year = "2024",
    address = "Miami, Florida, USA",
    publisher = "Association for Computational Linguistics",
    url = "https://aclanthology.org/2024.emnlp-main.64/",
    doi = "10.18653/v1/2024.emnlp-main.64",
    pages = "1107--1128"
}

@inproceedings{oktem-etal-2025-correcting,
    title = "Correcting the Tamazight Portions of {FLORES}+ and {OLDI} Seed Datasets",
    author = "Oktem, Alp  and
      Farhi, Mohamed Aymane  and
      Essaidi, Brahim  and
      Jabouja, Naceur  and
      Boudichat, Farida",
    editor = "Haddow, Barry  and
      Kocmi, Tom  and
      Koehn, Philipp  and
      Monz, Christof",
    booktitle = "Proceedings of the Tenth Conference on Machine Translation",
    month = nov,
    year = "2025",
    address = "Suzhou, China",
    publisher = "Association for Computational Linguistics",
    url = "https://aclanthology.org/2025.wmt-1.82/",
    pages = "1072--1080",
    ISBN = "979-8-89176-341-8"
}

@inproceedings{rajcoomar-2025-kozkreolmru,
    title = "{K}oz{K}reol{MRU} {WMT} 2025 {C}reole{MT} System Description: Koz Kreol: Multi-Stage Training for {E}nglish{--}Mauritian Creole {MT}",
    author = "Rajcoomar, Yush",
    editor = "Haddow, Barry  and
      Kocmi, Tom  and
      Koehn, Philipp  and
      Monz, Christof",
    booktitle = "Proceedings of the Tenth Conference on Machine Translation",
    month = nov,
    year = "2025",
    address = "Suzhou, China",
    publisher = "Association for Computational Linguistics",
    url = "https://aclanthology.org/2025.wmt-1.92/",
    doi = "10.18653/v1/2025.wmt-1.92",
    pages = "1183--1190",
    ISBN = "979-8-89176-341-8"
}

@inproceedings{frontull-etal-2025-bringing,
    title = "Bringing {L}adin to {FLORES}+",
    author = {Frontull, Samuel  and
      Str{\"o}hle, Thomas  and
      Zoli, Carlo  and
      Pescosta, Werner  and
      Frenademez, Ulrike  and
      Ruggeri, Matteo  and
      Valentin, Daria  and
      Comploj, Karin  and
      Perathoner, Gabriel  and
      Liotto, Silvia  and
      Anvidalfarei, Paolo},
    editor = "Haddow, Barry  and
      Kocmi, Tom  and
      Koehn, Philipp  and
      Monz, Christof",
    booktitle = "Proceedings of the Tenth Conference on Machine Translation",
    month = nov,
    year = "2025",
    address = "Suzhou, China",
    publisher = "Association for Computational Linguistics",
    url = "https://aclanthology.org/2025.wmt-1.81/",
    pages = "1061--1071",
    ISBN = "979-8-89176-341-8"
}

@misc{gemini25,
      title={Gemini 2.5: Pushing the Frontier with Advanced Reasoning, Multimodality, Long Context, and Next Generation Agentic Capabilities}, 
      author={Gheorghe Comanici and Eric Bieber and Mike Schaekermann and Ice Pasupat and Noveen Sachdeva and Inderjit Dhillon and Marcel Blistein and Ori Ram and Dan Zhang and Evan Rosen and Luke Marris and Sam Petulla and Colin Gaffney and Asaf Aharoni and Nathan Lintz and Tiago Cardal Pais and Henrik Jacobsson and Idan Szpektor and Nan-Jiang Jiang and Krishna Haridasan and Ahmed Omran and Nikunj Saunshi and Dara Bahri and Gaurav Mishra and Eric Chu and Toby Boyd and Brad Hekman and Aaron Parisi and Chaoyi Zhang and Kornraphop Kawintiranon and Tania Bedrax-Weiss and Oliver Wang and Ya Xu and Ollie Purkiss and Uri Mendlovic and Ilaï Deutel and Nam Nguyen and Adam Langley and Flip Korn and Lucia Rossazza and Alexandre Ramé and Sagar Waghmare and Helen Miller and Nathan Byrd and Ashrith Sheshan and Raia Hadsell and Sangnie Bhardwaj and Pawel Janus and Tero Rissa and Dan Horgan and Alvin Abdagic and Lior Belenki and James Allingham and Anima Singh and Theo Guidroz and Srivatsan Srinivasan and Herman Schmit and Kristen Chiafullo and Andre Elisseeff and Nilpa Jha and Prateek Kolhar and Leonard Berrada and Frank Ding and Xiance Si and Shrestha Basu Mallick and Franz Och and Sofia Erell and Eric Ni and Tejasi Latkar and Sherry Yang and Petar Sirkovic and Ziqiang Feng and Robert Leland and Rachel Hornung and Gang Wu and Charles Blundell and Hamidreza Alvari and Po-Sen Huang and Cathy Yip and Sanja Deur and Li Liu and Gabriela Surita and Pablo Duque and Dima Damen and Johnson Jia and Arthur Guez and Markus Mircea and Animesh Sinha and Alberto Magni and Paweł Stradomski and Tal Marian and Vlado Galić and Wenhu Chen and Hisham Husain and Achintya Singhal and Dominik Grewe and François-Xavier Aubet and Shuang Song and Lorenzo Blanco and Leland Rechis and Lewis Ho and Rich Munoz and Kelvin Zheng and Jessica Hamrick and Kevin Mather and Hagai Taitelbaum and Eliza Rutherford and Yun Lei and Kuangyuan Chen and Anand Shukla and Erica Moreira and Eric Doi and Berivan Isik and Nir Shabat and Dominika Rogozińska and Kashyap Kolipaka and Jason Chang and Eugen Vušak and Srinivasan Venkatachary and Shadi Noghabi and Tarun Bharti and Younghoon Jun and Aleksandr Zaks and Simon Green and Jeshwanth Challagundla and William Wong and Muqthar Mohammad and Dean Hirsch and Yong Cheng and Iftekhar Naim and Lev Proleev and Damien Vincent and Aayush Singh and Maxim Krikun and Dilip Krishnan and Zoubin Ghahramani and Aviel Atias and Rajeev Aggarwal and Christo Kirov and Dimitrios Vytiniotis and Christy Koh and Alexandra Chronopoulou and Pawan Dogra and Vlad-Doru Ion and Gladys Tyen and Jason Lee and Felix Weissenberger and Trevor Strohman and Ashwin Balakrishna and Jack Rae and Marko Velic and Raoul de Liedekerke and Oded Elyada and Wentao Yuan and Canoee Liu and Lior Shani and Sergey Kishchenko and Bea Alessio and Yandong Li and Richard Song and Sam Kwei and Orion Jankowski and Aneesh Pappu and Youhei Namiki and Yenai Ma and Nilesh Tripuraneni and Colin Cherry and Marissa Ikonomidis and Yu-Cheng Ling and Colin Ji and Beka Westberg and Auriel Wright and Da Yu and David Parkinson and Swaroop Ramaswamy and Jerome Connor and Soheil Hassas Yeganeh and Snchit Grover and George Kenwright and Lubo Litchev and Chris Apps and Alex Tomala and Felix Halim and Alex Castro-Ros and Zefei Li and Anudhyan Boral and Pauline Sho and Michal Yarom and Eric Malmi and David Klinghoffer and Rebecca Lin and Alan Ansell and Pradeep Kumar S and Shubin Zhao and Siqi Zuo and Adam Santoro and Heng-Tze Cheng and Solomon Demmessie and Yuchi Liu and Nicole Brichtova and Allie Culp and Nathaniel Braun and Dan Graur and Will Ng and Nikhil Mehta and Aaron Phillips and Patrik Sundberg and Varun Godbole and Fangyu Liu and Yash Katariya and David Rim and Mojtaba Seyedhosseini and Sean Ammirati and Jonas Valfridsson and Mahan Malihi and Timothy Knight and Andeep Toor and Thomas Lampe and Abe Ittycheriah and Lewis Chiang and Chak Yeung and Alexandre Fréchette and Jinmeng Rao and Huisheng Wang and Himanshu Srivastava and Richard Zhang and Rocky Rhodes and Ariel Brand and Dean Weesner and Ilya Figotin and Felix Gimeno and Rachana Fellinger and Pierre Marcenac and José Leal and Eyal Marcus and Victor Cotruta and Rodrigo Cabrera and Sheryl Luo and Dan Garrette and Vera Axelrod and Sorin Baltateanu and David Barker and Dongkai Chen and Horia Toma and Ben Ingram and Jason Riesa and Chinmay Kulkarni and Yujing Zhang and Hongbin Liu and Chao Wang and Martin Polacek and Will Wu and Kai Hui and Adrian N Reyes and Yi Su and Megan Barnes and Ishaan Malhi and Anfal Siddiqui and Qixuan Feng and Mihai Damaschin and Daniele Pighin and Andreas Steiner and Samuel Yang and Ramya Sree Boppana and Simeon Ivanov and Arun Kandoor and Aditya Shah and Asier Mujika and Da Huang and Christopher A. Choquette-Choo and Mohak Patel and Tianhe Yu and Toni Creswell and Jerry and Liu and Catarina Barros and Yasaman Razeghi and Aurko Roy and Phil Culliton and Binbin Xiong and Jiaqi Pan and Thomas Strohmann and Tolly Powell and Babi Seal and Doug DeCarlo and Pranav Shyam and Kaan Katircioglu and Xuezhi Wang and Cassidy Hardin and Immanuel Odisho and Josef Broder and Oscar Chang and Arun Nair and Artem Shtefan and Maura O'Brien and Manu Agarwal and Sahitya Potluri and Siddharth Goyal and Amit Jhindal and Saksham Thakur and Yury Stuken and James Lyon and Kristina Toutanova and Fangxiaoyu Feng and Austin Wu and Ben Horn and Alek Wang and Alex Cullum and Gabe Taubman and Disha Shrivastava and Chongyang Shi and Hamish Tomlinson and Roma Patel and Tao Tu and Ada Maksutaj Oflazer and Francesco Pongetti and Mingyao Yang and Adrien Ali Taïga and Vincent Perot and Nuo Wang Pierse and Feng Han and Yoel Drori and Iñaki Iturrate and Ayan Chakrabarti and Legg Yeung and Dave Dopson and Yi-ting Chen and Apoorv Kulshreshtha and Tongfei Guo and Philip Pham and Tal Schuster and Junquan Chen and Alex Polozov and Jinwei Xing and Huanjie Zhou and Praneeth Kacham and Doron Kukliansky and Antoine Miech and Sergey Yaroshenko and Ed Chi and Sholto Douglas and Hongliang Fei and Mathieu Blondel and Preethi Myla and Lior Madmoni and Xing Wu and Daniel Keysers and Kristian Kjems and Isabela Albuquerque and Lijun Yu and Joel D'sa and Michelle Plantan and Vlad Ionescu and Jaume Sanchez Elias and Abhirut Gupta and Manish Reddy Vuyyuru and Fred Alcober and Tong Zhou and Kaiyang Ji and Florian Hartmann and Subha Puttagunta and Hugo Song and Ehsan Amid and Anca Stefanoiu and Andrew Lee and Paul Pucciarelli and Emma Wang and Amit Raul and Slav Petrov and Isaac Tian and Valentin Anklin and Nana Nti and Victor Gomes and Max Schumacher and Grace Vesom and Alex Panagopoulos and Konstantinos Bousmalis and Daniel Andor and Josh Jacob and Yuan Zhang and Bill Rosgen and Matija Kecman and Matthew Tung and Alexandra Belias and Noah Goodman and Paul Covington and Brian Wieder and Nikita Saxena and Elnaz Davoodi and Muhuan Huang and Sharath Maddineni and Vincent Roulet and Folawiyo Campbell-Ajala and Pier Giuseppe Sessa and Xintian and Wu and Guangda Lai and Paul Collins and Alex Haig and Vytenis Sakenas and Xiaowei Xu and Marissa Giustina and Laurent El Shafey and Pichi Charoenpanit and Shefali Garg and Joshua Ainslie and Boone Severson and Montse Gonzalez Arenas and Shreya Pathak and Sujee Rajayogam and Jie Feng and Michiel Bakker and Sheng Li and Nevan Wichers and Jamie Rogers and Xinyang Geng and Yeqing Li and Rolf Jagerman and Chao Jia and Nadav Olmert and David Sharon and Matthew Mauger and Sandeep Mariserla and Hongxu Ma and Megha Mohabey and Kyuyeun Kim and Alek Andreev and Scott Pollom and Juliette Love and Vihan Jain and Priyanka Agrawal and Yannick Schroecker and Alisa Fortin and Manfred Warmuth and Ji Liu and Andrew Leach and Irina Blok and Ganesh Poomal Girirajan and Roee Aharoni and Benigno Uria and Andrei Sozanschi and Dan Goldberg and Lucian Ionita and Marco Tulio Ribeiro and Martin Zlocha and Vighnesh Birodkar and Sami Lachgar and Liangzhe Yuan and Himadri Choudhury and Matt Ginsberg and Fei Zheng and Gregory Dibb and Emily Graves and Swachhand Lokhande and Gabriel Rasskin and George-Cristian Muraru and Corbin Quick and Sandeep Tata and Pierre Sermanet and Aditya Chawla and Itay Karo and Yan Wang and Susan Zhang and Orgad Keller and Anca Dragan and Guolong Su and Ian Chou and Xi Liu and Yiqing Tao and Shruthi Prabhakara and Marc Wilson and Ruibo Liu and Shibo Wang and Georgie Evans and David Du and Alfonso Castaño and Gautam Prasad and Mona El Mahdy and Sebastian Gerlach and Machel Reid and Jarrod Kahn and Amir Zait and Thanumalayan Sankaranarayana Pillai and Thatcher Ulrich and Guanyu Wang and Jan Wassenberg and Efrat Farkash and Kiran Yalasangi and Congchao Wang and Maria Bauza and Simon Bucher and Ting Liu and Jun Yan and Gary Leung and Vikas Sindhwani and Parker Barnes and Avi Singh and Ivan Jurin and Jichuan Chang and Niket Kumar Bhumihar and Sivan Eiger and Gui Citovsky and Ben Withbroe and Zhang Li and Siyang Xue and Niccolò Dal Santo and Georgi Stoyanov and Yves Raimond and Steven Zheng and Yilin Gao and Vít Listík and Sławek Kwasiborski and Rachel Saputro and Adnan Ozturel and Ganesh Mallya and Kushal Majmundar and Ross West and Paul Caron and Jinliang Wei and Lluis Castrejon and Sharad Vikram and Deepak Ramachandran and Nikhil Dhawan and Jiho Park and Sara Smoot and George van den Driessche and Yochai Blau and Chase Malik and Wei Liang and Roy Hirsch and Cicero Nogueira dos Santos and Eugene Weinstein and Aäron van den Oord and Sid Lall and Nicholas FitzGerald and Zixuan Jiang and Xuan Yang and Dale Webster and Ali Elqursh and Aedan Pope and Georges Rotival and David Raposo and Wanzheng Zhu and Jeff Dean and Sami Alabed and Dustin Tran and Arushi Gupta and Zach Gleicher and Jessica Austin and Edouard Rosseel and Megh Umekar and Dipanjan Das and Yinghao Sun and Kai Chen and Karolis Misiunas and Xiang Zhou and Yixian Di and Alyssa Loo and Josh Newlan and Bo Li and Vinay Ramasesh and Ying Xu and Alex Chen and Sudeep Gandhe and Radu Soricut and Nikita Gupta and Shuguang Hu and Seliem El-Sayed and Xavier Garcia and Idan Brusilovsky and Pu-Chin Chen and Andrew Bolt and Lu Huang and Alex Gurney and Zhiying Zhang and Alexander Pritzel and Jarek Wilkiewicz and Bryan Seybold and Bhargav Kanagal Shamanna and Felix Fischer and Josef Dean and Karan Gill and Ross Mcilroy and Abhishek Bhowmick and Jeremy Selier and Antoine Yang and Derek Cheng and Vladimir Magay and Jie Tan and Dhriti Varma and Christian Walder and Tomas Kocisky and Ryo Nakashima and Paul Natsev and Mike Kwong and Ionel Gog and Chiyuan Zhang and Sander Dieleman and Thomas Jimma and Andrey Ryabtsev and Siddhartha Brahma and David Steiner and Dayou Du and Ante Žužul and Mislav Žanić and Mukund Raghavachari and Willi Gierke and Zeyu Zheng and Dessie Petrova and Yann Dauphin and Yuchuan Liu and Ido Kessler and Steven Hand and Chris Duvarney and Seokhwan Kim and Hyo Lee and Léonard Hussenot and Jeffrey Hui and Josh Smith and Deepali Jain and Jiawei Xia and Gaurav Singh Tomar and Keyvan Amiri and Du Phan and Fabian Fuchs and Tobias Weyand and Nenad Tomasev and Alexandra Cordell and Xin Liu and Jonathan Mallinson and Pankaj Joshi and Andy Crawford and Arun Suggala and Steve Chien and Nick Fernando and Mariella Sanchez-Vargas and Duncan Williams and Phil Crone and Xiyang Luo and Igor Karpov and Jyn Shan and Terry Thurk and Robin Strudel and Paul Voigtlaender and Piyush Patil and Tim Dozat and Ali Khodaei and Sahil Singla and Piotr Ambroszczyk and Qiyin Wu and Yifan Chang and Brian Roark and Chaitra Hegde and Tianli Ding and Angelos Filos and Zhongru Wu and André Susano Pinto and Shuang Liu and Saarthak Khanna and Aditya Pandey and Siobhan Mcloughlin and Qiujia Li and Sam Haves and Allan Zhou and Elena Buchatskaya and Isabel Leal and Peter de Boursac and Nami Akazawa and Nina Anderson and Terry Chen and Krishna Somandepalli and Chen Liang and Sheela Goenka and Stephanie Winkler and Alexander Grushetsky and Yifan Ding and Jamie Smith and Fan Ye and Jordi Pont-Tuset and Eric Li and Ruichao Li and Tomer Golany and Dawid Wegner and Tao Jiang and Omer Barak and Yuan Shangguan and Eszter Vértes and Renee Wong and Jörg Bornschein and Alex Tudor and Michele Bevilacqua and Tom Schaul and Ankit Singh Rawat and Yang Zhao and Kyriakos Axiotis and Lei Meng and Cory McLean and Jonathan Lai and Jennifer Beattie and Nate Kushman and Yaxin Liu and Blair Kutzman and Fiona Lang and Jingchen Ye and Praneeth Netrapalli and Pushkar Mishra and Myriam Khan and Megha Goel and Rob Willoughby and David Tian and Honglei Zhuang and JD Chen and Zak Tsai and Tasos Kementsietsidis and Arjun Khare and James Keeling and Keyang Xu and Nathan Waters and Florent Altché and Ashok Popat and Bhavishya Mittal and David Saxton and Dalia El Badawy and Michael Mathieu and Zheng Zheng and Hao Zhou and Nishant Ranka and Richard Shin and Qingnan Duan and Tim Salimans and Ioana Mihailescu and Uri Shaham and Ming-Wei Chang and Yannis Assael and Nishanth Dikkala and Martin Izzard and Vincent Cohen-Addad and Cat Graves and Vlad Feinberg and Grace Chung and DJ Strouse and Danny Karmon and Sahand Sharifzadeh and Zoe Ashwood and Khiem Pham and Jon Blanton and Alex Vasiloff and Jarred Barber and Mark Geller and Aurick Zhou and Fedir Zubach and Tzu-Kuo Huang and Lei Zhang and Himanshu Gupta and Matt Young and Julia Proskurnia and Ronny Votel and Valentin Gabeur and Gabriel Barcik and Aditya Tripathi and Hongkun Yu and Geng Yan and Beer Changpinyo and Filip Pavetić and Amy Coyle and Yasuhisa Fujii and Jorge Gonzalez Mendez and Tianhao Zhou and Harish Rajamani and Blake Hechtman and Eddie Cao and Da-Cheng Juan and Yi-Xuan Tan and Valentin Dalibard and Yilun Du and Natalie Clay and Kaisheng Yao and Wenhao Jia and Dimple Vijaykumar and Yuxiang Zhou and Xinyi Bai and Wei-Chih Hung and Steven Pecht and Georgi Todorov and Nikhil Khadke and Pramod Gupta and Preethi Lahoti and Arnaud Autef and Karthik Duddu and James Lee-Thorp and Alexander Bykovsky and Tautvydas Misiunas and Sebastian Flennerhag and Santhosh Thangaraj and Jed McGiffin and Zack Nado and Markus Kunesch and Andreas Noever and Amir Hertz and Marco Liang and Victor Stone and Evan Palmer and Samira Daruki and Arijit Pramanik and Siim Põder and Austin Kyker and Mina Khan and Evgeny Sluzhaev and Marvin Ritter and Avraham Ruderman and Wenlei Zhou and Chirag Nagpal and Kiran Vodrahalli and George Necula and Paul Barham and Ellie Pavlick and Jay Hartford and Izhak Shafran and Long Zhao and Maciej Mikuła and Tom Eccles and Hidetoshi Shimokawa and Kanav Garg and Luke Vilnis and Hanwen Chen and Ilia Shumailov and Kuang-Huei Lee and Abdelrahman Abdelhamed and Meiyan Xie and Vered Cohen and Ester Hlavnova and Dan Malkin and Chawin Sitawarin and James Lottes and Pauline Coquinot and Tianli Yu and Sandeep Kumar and Jingwei Zhang and Aroma Mahendru and Zafarali Ahmed and James Martens and Tao Chen and Aviel Boag and Daiyi Peng and Coline Devin and Arseniy Klimovskiy and Mary Phuong and Danny Vainstein and Jin Xie and Bhuvana Ramabhadran and Nathan Howard and Xinxin Yu and Gitartha Goswami and Jingyu Cui and Sam Shleifer and Mario Pinto and Chih-Kuan Yeh and Ming-Hsuan Yang and Sara Javanmardi and Dan Ethier and Chace Lee and Jordi Orbay and Suyog Kotecha and Carla Bromberg and Pete Shaw and James Thornton and Adi Gerzi Rosenthal and Shane Gu and Matt Thomas and Ian Gemp and Aditya Ayyar and Asahi Ushio and Aarush Selvan and Joel Wee and Chenxi Liu and Maryam Majzoubi and Weiren Yu and Jake Abernethy and Tyler Liechty and Renke Pan and Hoang Nguyen and Qiong and Hu and Sarah Perrin and Abhinav Arora and Emily Pitler and Weiyi Wang and Kaushik Shivakumar and Flavien Prost and Ben Limonchik and Jing Wang and Yi Gao and Timothee Cour and Shyamal Buch and Huan Gui and Maria Ivanova and Philipp Neubeck and Kelvin Chan and Lucy Kim and Huizhong Chen and Naman Goyal and Da-Woon Chung and Lu Liu and Yao Su and Anastasia Petrushkina and Jiajun Shen and Armand Joulin and Yuanzhong Xu and Stein Xudong Lin and Yana Kulizhskaya and Ciprian Chelba and Shobha Vasudevan and Eli Collins and Vasilisa Bashlovkina and Tony Lu and Doug Fritz and Jongbin Park and Yanqi Zhou and Chen Su and Richard Tanburn and Mikhail Sushkov and Mitchelle Rasquinha and Jinning Li and Jennifer Prendki and Yiming Li and Pallavi LV and Shriya Sharma and Hen Fitoussi and Hui Huang and Andrew Dai and Phuong Dao and Mike Burrows and Henry Prior and Danfeng Qin and Golan Pundak and Lars Lowe Sjoesund and Art Khurshudov and Zhenkai Zhu and Albert Webson and Elizabeth Kemp and Tat Tan and Saurabh Agrawal and Susie Sargsyan and Liqun Cheng and Jim Stephan and Tom Kwiatkowski and David Reid and Arunkumar Byravan and Assaf Hurwitz Michaely and Nicolas Heess and Luowei Zhou and Sonam Goenka and Viral Carpenter and Anselm Levskaya and Bo Wang and Reed Roberts and Rémi Leblond and Sharat Chikkerur and Stav Ginzburg and Max Chang and Robert Riachi and Chuqiao and Xu and Zalán Borsos and Michael Pliskin and Julia Pawar and Morgane Lustman and Hannah Kirkwood and Ankit Anand and Aditi Chaudhary and Norbert Kalb and Kieran Milan and Sean Augenstein and Anna Goldie and Laurel Prince and Karthik Raman and Yanhua Sun and Vivian Xia and Aaron Cohen and Zhouyuan Huo and Josh Camp and Seher Ellis and Lukas Zilka and David Vilar Torres and Lisa Patel and Sho Arora and Betty Chan and Jonas Adler and Kareem Ayoub and Jacky Liang and Fayaz Jamil and Jiepu Jiang and Simon Baumgartner and Haitian Sun and Yael Karov and Yaroslav Akulov and Hui Zheng and Irene Cai and Claudio Fantacci and James Rubin and Alex Rav Acha and Mengchao Wang and Nina D'Souza and Rohit Sathyanarayana and Shengyang Dai and Simon Rowe and Andrey Simanovsky and Omer Goldman and Yuheng Kuang and Xiaoyue Pan and Andrew Rosenberg and Tania Rojas-Esponda and Praneet Dutta and Amy Zeng and Irina Jurenka and Greg Farquhar and Yamini Bansal and Shariq Iqbal and Becca Roelofs and Ga-Young Joung and Parker Beak and Changwan Ryu and Ryan Poplin and Yan Wu and Jean-Baptiste Alayrac and Senaka Buthpitiya and Olaf Ronneberger and Caleb Habtegebriel and Wei Li and Paul Cavallaro and Aurora Wei and Guy Bensky and Timo Denk and Harish Ganapathy and Jeff Stanway and Pratik Joshi and Francesco Bertolini and Jessica Lo and Olivia Ma and Zachary Charles and Geta Sampemane and Himanshu Sahni and Xu Chen and Harry Askham and David Gaddy and Peter Young and Jiewen Tan and Matan Eyal and Arthur Bražinskas and Li Zhong and Zhichun Wu and Mark Epstein and Kai Bailey and Andrew Hard and Kamyu Lee and Sasha Goldshtein and Alex Ruiz and Mohammed Badawi and Matthias Lochbrunner and JK Kearns and Ashley Brown and Fabio Pardo and Theophane Weber and Haichuan Yang and Pan-Pan Jiang and Berkin Akin and Zhao Fu and Marcus Wainwright and Chi Zou and Meenu Gaba and Pierre-Antoine Manzagol and Wendy Kan and Yang Song and Karina Zainullina and Rui Lin and Jeongwoo Ko and Salil Deshmukh and Apoorv Jindal and James Svensson and Divya Tyam and Heri Zhao and Christine Kaeser-Chen and Scott Baird and Pooya Moradi and Jamie Hall and Qiuchen Guo and Vincent Tsang and Bowen Liang and Fernando Pereira and Suhas Ganesh and Ivan Korotkov and Jakub Adamek and Sridhar Thiagarajan and Vinh Tran and Charles Chen and Chris Tar and Sanil Jain and Ishita Dasgupta and Taylan Bilal and David Reitter and Kai Zhao and Giulia Vezzani and Yasmin Gehman and Pulkit Mehta and Lauren Beltrone and Xerxes Dotiwalla and Sergio Guadarrama and Zaheer Abbas and Stefani Karp and Petko Georgiev and Chun-Sung Ferng and Marc Brockschmidt and Liqian Peng and Christoph Hirnschall and Vikas Verma and Yingying Bi and Ying Xiao and Avigail Dabush and Kelvin Xu and Phil Wallis and Randall Parker and Qifei Wang and Yang Xu and Ilkin Safarli and Dinesh Tewari and Yin Zhang and Seungyeon Kim and Andrea Gesmundo and Mackenzie Thomas and Sergey Levi and Ahmed Chowdhury and Kanishka Rao and Peter Garst and Sam Conway-Rahman and Helen Ran and Kay McKinney and Zhisheng Xiao and Wenhao Yu and Rohan Agrawal and Axel Stjerngren and Catalin Ionescu and Jingjing Chen and Vivek Sharma and Justin Chiu and Fei Liu and Ken Franko and Clayton Sanford and Xingyu Cai and Paul Michel and Sanjay Ganapathy and Jane Labanowski and Zachary Garrett and Ben Vargas and Sean Sun and Bryan Gale and Thomas Buschmann and Guillaume Desjardins and Nimesh Ghelani and Palak Jain and Mudit Verma and Chulayuth Asawaroengchai and Julian Eisenschlos and Jitendra Harlalka and Hideto Kazawa and Don Metzler and Joshua Howland and Ying Jian and Jake Ades and Viral Shah and Tynan Gangwani and Seungji Lee and Roman Ring and Steven M. Hernandez and Dean Reich and Amer Sinha and Ashutosh Sathe and Joe Kovac and Ashleah Gill and Ajay Kannan and Andrea D'olimpio and Martin Sevenich and Jay Whang and Been Kim and Khe Chai Sim and Jilin Chen and Jiageng Zhang and Shuba Lall and Yossi Matias and Bill Jia and Abe Friesen and Sara Nasso and Ashish Thapliyal and Bryan Perozzi and Ting Yu and Anna Shekhawat and Safeen Huda and Peter Grabowski and Eric Wang and Ashwin Sreevatsa and Hilal Dib and Mehadi Hassen and Parker Schuh and Vedrana Milutinovic and Chris Welty and Michael Quinn and Ali Shah and Bangju Wang and Gabe Barth-Maron and Justin Frye and Natalie Axelsson and Tao Zhu and Yukun Ma and Irene Giannoumis and Hanie Sedghi and Chang Ye and Yi Luan and Kevin Aydin and Bilva Chandra and Vivek Sampathkumar and Ronny Huang and Victor Lavrenko and Ahmed Eleryan and Zhi Hong and Steven Hansen and Sara Mc Carthy and Bidisha Samanta and Domagoj Ćevid and Xin Wang and Fangtao Li and Michael Voznesensky and Matt Hoffman and Andreas Terzis and Vikash Sehwag and Gil Fidel and Luheng He and Mu Cai and Yanzhang He and Alex Feng and Martin Nikoltchev and Samrat Phatale and Jason Chase and Rory Lawton and Ming Zhang and Tom Ouyang and Manuel Tragut and Mehdi Hafezi Manshadi and Arjun Narayanan and Jiaming Shen and Xu Gao and Tolga Bolukbasi and Nick Roy and Xin Li and Daniel Golovin and Liviu Panait and Zhen Qin and Guangxing Han and Thomas Anthony and Sneha Kudugunta and Viorica Patraucean and Aniket Ray and Xinyun Chen and Xiaochen Yang and Tanuj Bhatia and Pranav Talluri and Alex Morris and Andrija Ražnatović and Bethanie Brownfield and James An and Sheng Peng and Patrick Kane and Ce Zheng and Nico Duduta and Joshua Kessinger and James Noraky and Siqi Liu and Keran Rong and Petar Veličković and Keith Rush and Alex Goldin and Fanny Wei and Shiva Mohan Reddy Garlapati and Caroline Pantofaru and Okwan Kwon and Jianmo Ni and Eric Noland and Julia Di Trapani and Françoise Beaufays and Abhijit Guha Roy and Yinlam Chow and Aybuke Turker and Geoffrey Cideron and Lantao Mei and Jon Clark and Qingyun Dou and Matko Bošnjak and Ralph Leith and Yuqing Du and Amir Yazdanbakhsh and Milad Nasr and Chester Kwak and Suraj Satishkumar Sheth and Alex Kaskasoli and Ankesh Anand and Balaji Lakshminarayanan and Sammy Jerome and David Bieber and Chun-Te Chu and Alexandre Senges and Tianxiao Shen and Mukund Sridhar and Ndaba Ndebele and Benjamin Beyret and Shakir Mohamed and Mia Chen and Markus Freitag and Jiaxian Guo and Luyang Liu and Paul Roit and Heng Chen and Shen Yan and Tom Stone and JD Co-Reyes and Jeremy Cole and Salvatore Scellato and Shekoofeh Azizi and Hadi Hashemi and Alicia Jin and Anand Iyer and Marcella Valentine and András György and Arun Ahuja and Daniel Hernandez Diaz and Chen-Yu Lee and Nathan Clement and Weize Kong and Drew Garmon and Ishaan Watts and Kush Bhatia and Khyatti Gupta and Matt Miecnikowski and Hugo Vallet and Ankur Taly and Edward Loper and Saket Joshi and James Atwood and Jo Chick and Mark Collier and Fotis Iliopoulos and Ryan Trostle and Beliz Gunel and Ramiro Leal-Cavazos and Arnar Mar Hrafnkelsson and Michael Guzman and Xiaoen Ju and Andy Forbes and Jesse Emond and Kushal Chauhan and Ben Caine and Li Xiao and Wenjun Zeng and Alexandre Moufarek and Daniel Murphy and Maya Meng and Nitish Gupta and Felix Riedel and Anil Das and Elijah Lawal and Shashi Narayan and Tiberiu Sosea and James Swirhun and Linda Friso and Behnam Neyshabur and Jing Lu and Sertan Girgin and Michael Wunder and Edouard Yvinec and Aroonalok Pyne and Victor Carbune and Shruti Rijhwani and Yang Guo and Tulsee Doshi and Anton Briukhov and Max Bain and Ayal Hitron and Xuanhui Wang and Ashish Gupta and Ke Chen and Cosmo Du and Weiyang Zhang and Dhruv Shah and Arjun Akula and Max Dylla and Ashyana Kachra and Weicheng Kuo and Tingting Zou and Lily Wang and Luyao Xu and Jifan Zhu and Justin Snyder and Sachit Menon and Orhan Firat and Igor Mordatch and Yuan Yuan and Natalia Ponomareva and Rory Blevins and Lawrence Moore and Weijun Wang and Phil Chen and Martin Scholz and Artur Dwornik and Jason Lin and Sicheng Li and Diego Antognini and Te I and Xiaodan Song and Matt Miller and Uday Kalra and Adam Raveret and Oscar Akerlund and Felix Wu and Andrew Nystrom and Namrata Godbole and Tianqi Liu and Hannah DeBalsi and Jewel Zhao and Buhuang Liu and Avi Caciularu and Lauren Lax and Urvashi Khandelwal and Victoria Langston and Eric Bailey and Silvio Lattanzi and Yufei Wang and Neel Kovelamudi and Sneha Mondal and Guru Guruganesh and Nan Hua and Ofir Roval and Paweł Wesołowski and Rishikesh Ingale and Jonathan Halcrow and Tim Sohn and Christof Angermueller and Bahram Raad and Eli Stickgold and Eva Lu and Alec Kosik and Jing Xie and Timothy Lillicrap and Austin Huang and Lydia Lihui Zhang and Dominik Paulus and Clement Farabet and Alex Wertheim and Bing Wang and Rishabh Joshi and Chu-ling Ko and Yonghui Wu and Shubham Agrawal and Lily Lin and XiangHai Sheng and Peter Sung and Tyler Breland-King and Christina Butterfield and Swapnil Gawde and Sumeet Singh and Qiao Zhang and Raj Apte and Shilpa Shetty and Adrian Hutter and Tao Li and Elizabeth Salesky and Federico Lebron and Jonni Kanerva and Michela Paganini and Arthur Nguyen and Rohith Vallu and Jan-Thorsten Peter and Sarmishta Velury and David Kao and Jay Hoover and Anna Bortsova and Colton Bishop and Shoshana Jakobovits and Alessandro Agostini and Alekh Agarwal and Chang Liu and Charles Kwong and Sasan Tavakkol and Ioana Bica and Alex Greve and Anirudh GP and Jake Marcus and Le Hou and Tom Duerig and Rivka Moroshko and Dave Lacey and Andy Davis and Julien Amelot and Guohui Wang and Frank Kim and Theofilos Strinopoulos and Hui Wan and Charline Le Lan and Shankar Krishnan and Haotian Tang and Peter Humphreys and Junwen Bai and Idan Heimlich Shtacher and Diego Machado and Chenxi Pang and Ken Burke and Dangyi Liu and Renga Aravamudhan and Yue Song and Ed Hirst and Abhimanyu Singh and Brendan Jou and Liang Bai and Francesco Piccinno and Chuyuan Kelly Fu and Robin Alazard and Barak Meiri and Daniel Winter and Charlie Chen and Mingda Zhang and Jens Heitkaemper and John Lambert and Jinhyuk Lee and Alexander Frömmgen and Sergey Rogulenko and Pranav Nair and Paul Niemczyk and Anton Bulyenov and Bibo Xu and Hadar Shemtov and Morteza Zadimoghaddam and Serge Toropov and Mateo Wirth and Hanjun Dai and Sreenivas Gollapudi and Daniel Zheng and Alex Kurakin and Chansoo Lee and Kalesha Bullard and Nicolas Serrano and Ivana Balazevic and Yang Li and Johan Schalkwyk and Mark Murphy and Mingyang Zhang and Kevin Sequeira and Romina Datta and Nishant Agrawal and Charles Sutton and Nithya Attaluri and Mencher Chiang and Wael Farhan and Gregory Thornton and Kate Lin and Travis Choma and Hung Nguyen and Kingshuk Dasgupta and Dirk Robinson and Iulia Comşa and Michael Riley and Arjun Pillai and Basil Mustafa and Ben Golan and Amir Zandieh and Jean-Baptiste Lespiau and Billy Porter and David Ross and Sujeevan Rajayogam and Mohit Agarwal and Subhashini Venugopalan and Bobak Shahriari and Qiqi Yan and Hao Xu and Taylor Tobin and Pavel Dubov and Hongzhi Shi and Adrià Recasens and Anton Kovsharov and Sebastian Borgeaud and Lucio Dery and Shanthal Vasanth and Elena Gribovskaya and Linhai Qiu and Mahdis Mahdieh and Wojtek Skut and Elizabeth Nielsen and CJ Zheng and Adams Yu and Carrie Grimes Bostock and Shaleen Gupta and Aaron Archer and Chris Rawles and Elinor Davies and Alexey Svyatkovskiy and Tomy Tsai and Yoni Halpern and Christian Reisswig and Bartek Wydrowski and Bo Chang and Joan Puigcerver and Mor Hazan Taege and Jian Li and Eva Schnider and Xinjian Li and Dragos Dena and Yunhan Xu and Umesh Telang and Tianze Shi and Heiga Zen and Kyle Kastner and Yeongil Ko and Neesha Subramaniam and Aviral Kumar and Pete Blois and Zhuyun Dai and John Wieting and Yifeng Lu and Yoel Zeldes and Tian Xie and Anja Hauth and Alexandru Ţifrea and Yuqi Li and Sam El-Husseini and Dan Abolafia and Howard Zhou and Wen Ding and Sahra Ghalebikesabi and Carlos Guía and Andrii Maksai and Ágoston Weisz and Sercan Arik and Nick Sukhanov and Aga Świetlik and Xuhui Jia and Luo Yu and Weiyue Wang and Mark Brand and Dawn Bloxwich and Sean Kirmani and Zhe Chen and Alec Go and Pablo Sprechmann and Nithish Kannen and Alen Carin and Paramjit Sandhu and Isabel Edkins and Leslie Nooteboom and Jai Gupta and Loren Maggiore and Javad Azizi and Yael Pritch and Pengcheng Yin and Mansi Gupta and Danny Tarlow and Duncan Smith and Desi Ivanov and Mohammad Babaeizadeh and Ankita Goel and Satish Kambala and Grace Chu and Matej Kastelic and Michelle Liu and Hagen Soltau and Austin Stone and Shivani Agrawal and Min Kim and Kedar Soparkar and Srinivas Tadepalli and Oskar Bunyan and Rachel Soh and Arvind Kannan and DY Kim and Blake JianHang Chen and Afief Halumi and Sudeshna Roy and Yulong Wang and Olcan Sercinoglu and Gena Gibson and Sijal Bhatnagar and Motoki Sano and Daniel von Dincklage and Qingchun Ren and Blagoj Mitrevski and Mirek Olšák and Jennifer She and Carl Doersch and Jilei and Wang and Bingyuan Liu and Qijun Tan and Tamar Yakar and Tris Warkentin and Alex Ramirez and Carl Lebsack and Josh Dillon and Rajiv Mathews and Tom Cobley and Zelin Wu and Zhuoyuan Chen and Jon Simon and Swaroop Nath and Tara Sainath and Alexei Bendebury and Ryan Julian and Bharath Mankalale and Daria Ćurko and Paulo Zacchello and Adam R. Brown and Kiranbir Sodhia and Heidi Howard and Sergi Caelles and Abhinav Gupta and Gareth Evans and Anna Bulanova and Lesley Katzen and Roman Goldenberg and Anton Tsitsulin and Joe Stanton and Benoit Schillings and Vitaly Kovalev and Corey Fry and Rushin Shah and Kuo Lin and Shyam Upadhyay and Cheng Li and Soroush Radpour and Marcello Maggioni and Jing Xiong and Lukas Haas and Jenny Brennan and Aishwarya Kamath and Nikolay Savinov and Arsha Nagrani and Trevor Yacovone and Ryan Kappedal and Kostas Andriopoulos and Li Lao and YaGuang Li and Grigory Rozhdestvenskiy and Kazuma Hashimoto and Andrew Audibert and Sophia Austin and Daniel Rodriguez and Anian Ruoss and Garrett Honke and Deep Karkhanis and Xi Xiong and Qing Wei and James Huang and Zhaoqi Leng and Vittal Premachandran and Stan Bileschi and Georgios Evangelopoulos and Thomas Mensink and Jay Pavagadhi and Denis Teplyashin and Paul Chang and Linting Xue and Garrett Tanzer and Sally Goldman and Kaushal Patel and Shixin Li and Jeremy Wiesner and Ivy Zheng and Ian Stewart-Binks and Jie Han and Zhi Li and Liangchen Luo and Karel Lenc and Mario Lučić and Fuzhao Xue and Ryan Mullins and Alexey Guseynov and Chung-Ching Chang and Isaac Galatzer-Levy and Adam Zhang and Garrett Bingham and Grace Hu and Ale Hartman and Yue Ma and Jordan Griffith and Alex Irpan and Carey Radebaugh and Summer Yue and Lijie Fan and Victor Ungureanu and Christina Sorokin and Hannah Teufel and Peiran Li and Rohan Anil and Dimitris Paparas and Todd Wang and Chu-Cheng Lin and Hui Peng and Megan Shum and Goran Petrovic and Demetra Brady and Richard Nguyen and Klaus Macherey and Zhihao Li and Harman Singh and Madhavi Yenugula and Mariko Iinuma and Xinyi Chen and Kavya Kopparapu and Alexey Stern and Shachi Dave and Chandu Thekkath and Florence Perot and Anurag Kumar and Fangda Li and Yang Xiao and Matthew Bilotti and Mohammad Hossein Bateni and Isaac Noble and Lisa Lee and Amelio Vázquez-Reina and Julian Salazar and Xiaomeng Yang and Boyu Wang and Ela Gruzewska and Anand Rao and Sindhu Raghuram and Zheng Xu and Eyal Ben-David and Jieru Mei and Sid Dalmia and Zhaoyi Zhang and Yuchen Liu and Gagan Bansal and Helena Pankov and Steven Schwarcz and Andrea Burns and Christine Chan and Sumit Sanghai and Ricky Liang and Ethan Liang and Antoine He and Amy Stuart and Arun Narayanan and Yukun Zhu and Christian Frank and Bahar Fatemi and Amit Sabne and Oran Lang and Indro Bhattacharya and Shane Settle and Maria Wang and Brendan McMahan and Andrea Tacchetti and Livio Baldini Soares and Majid Hadian and Serkan Cabi and Timothy Chung and Nikita Putikhin and Gang Li and Jeremy Chen and Austin Tarango and Henryk Michalewski and Mehran Kazemi and Hussain Masoom and Hila Sheftel and Rakesh Shivanna and Archita Vadali and Ramona Comanescu and Doug Reid and Joss Moore and Arvind Neelakantan and Michaël Sander and Jonathan Herzig and Aviv Rosenberg and Mostafa Dehghani and JD Choi and Michael Fink and Reid Hayes and Eric Ge and Shitao Weng and Chia-Hua Ho and John Karro and Kalpesh Krishna and Lam Nguyen Thiet and Amy Skerry-Ryan and Daniel Eppens and Marco Andreetto and Navin Sarma and Silvano Bonacina and Burcu Karagol Ayan and Megha Nawhal and Zhihao Shan and Mike Dusenberry and Shantanu Thakoor and Sagar Gubbi and Duc Dung Nguyen and Reut Tsarfaty and Samuel Albanie and Jovana Mitrović and Meet Gandhi and Bo-Juen Chen and Alessandro Epasto and Georgi Stephanov and Ye Jin and Samuel Gehman and Aida Amini and Jack Weber and Feryal Behbahani and Shawn Xu and Miltos Allamanis and Xi Chen and Myle Ott and Claire Sha and Michal Jastrzebski and Hang Qi and David Greene and Xinyi Wu and Abodunrinwa Toki and Daniel Vlasic and Jane Shapiro and Ragha Kotikalapudi and Zhe Shen and Takaaki Saeki and Sirui Xie and Albin Cassirer and Shikhar Bharadwaj and Tatsuya Kiyono and Srinadh Bhojanapalli and Elan Rosenfeld and Sam Ritter and Jieming Mao and João Gabriel Oliveira and Zoltan Egyed and Bernd Bandemer and Emilio Parisotto and Keisuke Kinoshita and Juliette Pluto and Petros Maniatis and Steve Li and Yaohui Guo and Golnaz Ghiasi and Jean Tarbouriech and Srimon Chatterjee and Julie Jin and Katrina and Xu and Jennimaria Palomaki and Séb Arnold and Madhavi Sewak and Federico Piccinini and Mohit Sharma and Ben Albrecht and Sean Purser-haskell and Ashwin Vaswani and Chongyan Chen and Matheus Wisniewski and Qin Cao and John Aslanides and Nguyet Minh Phu and Maximilian Sieb and Lauren Agubuzu and Anne Zheng and Daniel Sohn and Marco Selvi and Anders Andreassen and Krishan Subudhi and Prem Eruvbetine and Oliver Woodman and Tomas Mery and Sebastian Krause and Xiaoqi Ren and Xiao Ma and Jincheng Luo and Dawn Chen and Wei Fan and Henry Griffiths and Christian Schuler and Alice Li and Shujian Zhang and Jean-Michel Sarr and Shixin Luo and Riccardo Patana and Matthew Watson and Dani Naboulsi and Michael Collins and Sailesh Sidhwani and Emiel Hoogeboom and Sharon Silver and Emily Caveness and Xiaokai Zhao and Mikel Rodriguez and Maxine Deines and Libin Bai and Patrick Griffin and Marco Tagliasacchi and Emily Xue and Spandana Raj Babbula and Bo Pang and Nan Ding and Gloria Shen and Elijah Peake and Remi Crocker and Shubha Srinivas Raghvendra and Danny Swisher and Woohyun Han and Richa Singh and Ling Wu and Vladimir Pchelin and Tsendsuren Munkhdalai and Dana Alon and Geoff Bacon and Efren Robles and Jannis Bulian and Melvin Johnson and George Powell and Felipe Tiengo Ferreira and Yaoyiran Li and Frederik Benzing and Mihajlo Velimirović and Hubert Soyer and William Kong and Tony and Nguyên and Zhen Yang and Jeremiah Liu and Joost van Amersfoort and Daniel Gillick and Baochen Sun and Nathalie Rauschmayr and Katie Zhang and Serena Zhan and Tao Zhou and Alexey Frolov and Chengrun Yang and Denis Vnukov and Louis Rouillard and Hongji Li and Amol Mandhane and Nova Fallen and Rajesh Venkataraman and Clara Huiyi Hu and Jennifer Brennan and Jenny Lee and Jerry Chang and Martin Sundermeyer and Zhufeng Pan and Rosemary Ke and Simon Tong and Alex Fabrikant and William Bono and Jindong Gu and Ryan Foley and Yiran Mao and Manolis Delakis and Dhruva Bhaswar and Roy Frostig and Nick Li and Avital Zipori and Cath Hope and Olga Kozlova and Swaroop Mishra and Josip Djolonga and Craig Schiff and Majd Al Merey and Eleftheria Briakou and Peter Morgan and Andy Wan and Avinatan Hassidim and RJ Skerry-Ryan and Kuntal Sengupta and Mary Jasarevic and Praveen Kallakuri and Paige Kunkle and Hannah Brennan and Tom Lieber and Hassan Mansoor and Julian Walker and Bing Zhang and Annie Xie and Goran Žužić and Adaeze Chukwuka and Alex Druinsky and Donghyun Cho and Rui Yao and Ferjad Naeem and Shiraz Butt and Eunyoung Kim and Zhipeng Jia and Mandy Jordan and Adam Lelkes and Mark Kurzeja and Sophie Wang and James Zhao and Andrew Over and Abhishek Chakladar and Marcel Prasetya and Neha Jha and Sriram Ganapathy and Yale Cong and Prakash Shroff and Carl Saroufim and Sobhan Miryoosefi and Mohamed Hammad and Tajwar Nasir and Weijuan Xi and Yang Gao and Young Maeng and Ben Hora and Chin-Yi Cheng and Parisa Haghani and Yoad Lewenberg and Caden Lu and Martin Matysiak and Naina Raisinghani and Huiyu Wang and Lexi Baugher and Rahul Sukthankar and Minh Giang and John Schultz and Noah Fiedel and Minmin Chen and Cheng-Chun Lee and Tapomay Dey and Hao Zheng and Shachi Paul and Celine Smith and Andy Ly and Yicheng Wang and Rishabh Bansal and Bartek Perz and Susanna Ricco and Stasha Blank and Vaishakh Keshava and Deepak Sharma and Marvin Chow and Kunal Lad and Komal Jalan and Simon Osindero and Craig Swanson and Jacob Scott and Anastasija Ilić and Xiaowei Li and Siddhartha Reddy Jonnalagadda and Afzal Shama Soudagar and Yan Xiong and Bat-Orgil Batsaikhan and Daniel Jarrett and Naveen Kumar and Maulik Shah and Matt Lawlor and Austin Waters and Mark Graham and Rhys May and Sabela Ramos and Sandra Lefdal and Zeynep Cankara and Nacho Cano and Brendan O'Donoghue and Jed Borovik and Frederick Liu and Jordan Grimstad and Mahmoud Alnahlawi and Katerina Tsihlas and Tom Hudson and Nikolai Grigorev and Yiling Jia and Terry Huang and Tobenna Peter Igwe and Sergei Lebedev and Xiaodan Tang and Igor Krivokon and Frankie Garcia and Melissa Tan and Eric Jia and Peter Stys and Shikhar Vashishth and Yu Liang and Balaji Venkatraman and Chenjie Gu and Anastasios Kementsietsidis and Chen Zhu and Junehyuk Jung and Yunfei Bai and Mohammad Javad Hosseini and Faruk Ahmed and Aditya Gupta and Xin Yuan and Shereen Ashraf and Shitij Nigam and Gautam Vasudevan and Pranjal Awasthi and Adi Mayrav Gilady and Zelda Mariet and Ramy Eskander and Haiguang Li and Hexiang Hu and Guillermo Garrido and Philippe Schlattner and George Zhang and Rohun Saxena and Petar Dević and Kritika Muralidharan and Ashwin Murthy and Yiqian Zhou and Min Choi and Arissa Wongpanich and Zhengdong Wang and Premal Shah and Yuntao Xu and Yiling Huang and Stephen Spencer and Alice Chen and James Cohan and Junjie Wang and Jonathan Tompson and Junru Wu and Ruba Haroun and Haiqiong Li and Blanca Huergo and Fan Yang and Tongxin Yin and James Wendt and Michael Bendersky and Rahma Chaabouni and Javier Snaider and Johan Ferret and Abhishek Jindal and Tara Thompson and Andrew Xue and Will Bishop and Shubham Milind Phal and Archit Sharma and Yunhsuan Sung and Prabakar Radhakrishnan and Mo Shomrat and Reeve Ingle and Roopali Vij and Justin Gilmer and Mihai Dorin Istin and Sam Sobell and Yang Lu and Emily Nottage and Dorsa Sadigh and Jeremiah Willcock and Tingnan Zhang and Steve Xu and Sasha Brown and Katherine Lee and Gary Wang and Yun Zhu and Yi Tay and Cheolmin Kim and Audrey Gutierrez and Abhanshu Sharma and Yongqin Xian and Sungyong Seo and Claire Cui and Elena Pochernina and Cip Baetu and Krzysztof Jastrzębski and Mimi Ly and Mohamed Elhawaty and Dan Suh and Eren Sezener and Pidong Wang and Nancy Yuen and George Tucker and Jiahao Cai and Zuguang Yang and Cindy Wang and Alex Muzio and Hai Qian and Jae Yoo and Derek Lockhart and Kevin R. McKee and Mandy Guo and Malika Mehrotra and Artur Mendonça and Sanket Vaibhav Mehta and Sherry Ben and Chetan Tekur and Jiaqi Mu and Muye Zhu and Victoria Krakovna and Hongrae Lee and AJ Maschinot and Sébastien Cevey and HyunJeong Choe and Aijun Bai and Hansa Srinivasan and Derek Gasaway and Nick Young and Patrick Siegler and Dan Holtmann-Rice and Vihari Piratla and Kate Baumli and Roey Yogev and Alex Hofer and Hado van Hasselt and Svetlana Grant and Yuri Chervonyi and David Silver and Andrew Hogue and Ayushi Agarwal and Kathie Wang and Preeti Singh and Four Flynn and Josh Lipschultz and Robert David and Lizzetth Bellot and Yao-Yuan Yang and Long Le and Filippo Graziano and Kate Olszewska and Kevin Hui and Akanksha Maurya and Nikos Parotsidis and Weijie Chen and Tayo Oguntebi and Joe Kelley and Anirudh Baddepudi and Johannes Mauerer and Gregory Shaw and Alex Siegman and Lin Yang and Shravya Shetty and Subhrajit Roy and Yunting Song and Wojciech Stokowiec and Ryan Burnell and Omkar Savant and Robert Busa-Fekete and Jin Miao and Samrat Ghosh and Liam MacDermed and Phillip Lippe and Mikhail Dektiarev and Zach Behrman and Fabian Mentzer and Kelvin Nguyen and Meng Wei and Siddharth Verma and Chris Knutsen and Sudeep Dasari and Zhipeng Yan and Petr Mitrichev and Xingyu Wang and Virat Shejwalkar and Jacob Austin and Srinivas Sunkara and Navneet Potti and Yan Virin and Christian Wright and Gaël Liu and Oriana Riva and Etienne Pot and Greg Kochanski and Quoc Le and Gargi Balasubramaniam and Arka Dhar and Yuguo Liao and Adam Bloniarz and Divyansh Shukla and Elizabeth Cole and Jong Lee and Sheng Zhang and Sushant Kafle and Siddharth Vashishtha and Parsa Mahmoudieh and Grace Chen and Raphael Hoffmann and Pranesh Srinivasan and Agustin Dal Lago and Yoav Ben Shalom and Zi Wang and Michael Elabd and Anuj Sharma and Junhyuk Oh and Suraj Kothawade and Maigo Le and Marianne Monteiro and Shentao Yang and Kaiz Alarakyia and Robert Geirhos and Diana Mincu and Håvard Garnes and Hayato Kobayashi and Soroosh Mariooryad and Kacper Krasowiak and Zhixin and Lai and Shibl Mourad and Mingqiu Wang and Fan Bu and Ophir Aharoni and Guanjie Chen and Abhimanyu Goyal and Vadim Zubov and Ankur Bapna and Elahe Dabir and Nisarg Kothari and Kay Lamerigts and Nicola De Cao and Jeremy Shar and Christopher Yew and Nitish Kulkarni and Dre Mahaarachchi and Mandar Joshi and Zhenhai Zhu and Jared Lichtarge and Yichao Zhou and Hannah Muckenhirn and Vittorio Selo and Oriol Vinyals and Peter Chen and Anthony Brohan and Vaibhav Mehta and Sarah Cogan and Ruth Wang and Ty Geri and Wei-Jen Ko and Wei Chen and Fabio Viola and Keshav Shivam and Lisa Wang and Madeleine Clare Elish and Raluca Ada Popa and Sébastien Pereira and Jianqiao Liu and Raphael Koster and Donnie Kim and Gufeng Zhang and Sayna Ebrahimi and Partha Talukdar and Yanyan Zheng and Petra Poklukar and Ales Mikhalap and Dale Johnson and Anitha Vijayakumar and Mark Omernick and Matt Dibb and Ayush Dubey and Qiong Hu and Apurv Suman and Vaibhav Aggarwal and Ilya Kornakov and Fei Xia and Wing Lowe and Alexey Kolganov and Ted Xiao and Vitaly Nikolaev and Steven Hemingray and Bonnie Li and Joana Iljazi and Mikołaj Rybiński and Ballie Sandhu and Peggy Lu and Thang Luong and Rodolphe Jenatton and Vineetha Govindaraj and Hui and Li and Gabriel Dulac-Arnold and Wonpyo Park and Henry Wang and Abhinit Modi and Jean Pouget-Abadie and Kristina Greller and Rahul Gupta and Robert Berry and Prajit Ramachandran and Jinyu Xie and Liam McCafferty and Jianling Wang and Kilol Gupta and Hyeontaek Lim and Blaž Bratanič and Andy Brock and Ilia Akolzin and Jim Sproch and Dan Karliner and Duhyeon Kim and Adrian Goedeckemeyer and Noam Shazeer and Cordelia Schmid and Daniele Calandriello and Parul Bhatia and Krzysztof Choromanski and Ceslee Montgomery and Dheeru Dua and Ana Ramalho and Helen King and Yue Gao and Lynn Nguyen and David Lindner and Divya Pitta and Oleaser Johnson and Khalid Salama and Diego Ardila and Michael Han and Erin Farnese and Seth Odoom and Ziyue Wang and Xiangzhuo Ding and Norman Rink and Ray Smith and Harshal Tushar Lehri and Eden Cohen and Neera Vats and Tong He and Parthasarathy Gopavarapu and Adam Paszke and Miteyan Patel and Wouter Van Gansbeke and Lucia Loher and Luis Castro and Maria Voitovich and Tamara von Glehn and Nelson George and Simon Niklaus and Zach Eaton-Rosen and Nemanja Rakićević and Erik Jue and Sagi Perel and Carrie Zhang and Yuval Bahat and Angéline Pouget and Zhi Xing and Fantine Huot and Ashish Shenoy and Taylor Bos and Vincent Coriou and Bryan Richter and Natasha Noy and Yaqing Wang and Santiago Ontanon and Siyang Qin and Gleb Makarchuk and Demis Hassabis and Zhuowan Li and Mandar Sharma and Kumaran Venkatesan and Iurii Kemaev and Roxanne Daniel and Shiyu Huang and Saloni Shah and Octavio Ponce and Warren and Chen and Manaal Faruqui and Jialin Wu and Slavica Andačić and Szabolcs Payrits and Daniel McDuff and Tom Hume and Yuan Cao and MH Tessler and Qingze Wang and Yinan Wang and Ivor Rendulic and Eirikur Agustsson and Matthew Johnson and Tanya Lando and Andrew Howard and Sri Gayatri Sundara Padmanabhan and Mayank Daswani and Andrea Banino and Michael Kilgore and Jonathan Heek and Ziwei Ji and Alvaro Caceres and Conglong Li and Nora Kassner and Alexey Vlaskin and Zeyu Liu and Alex Grills and Yanhan Hou and Roykrong Sukkerd and Gowoon Cheon and Nishita Shetty and Larisa Markeeva and Piotr Stanczyk and Tejas Iyer and Yuan Gong and Shawn Gao and Keerthana Gopalakrishnan and Tim Blyth and Malcolm Reynolds and Avishkar Bhoopchand and Misha Bilenko and Dero Gharibian and Vicky Zayats and Aleksandra Faust and Abhinav Singh and Min Ma and Hongyang Jiao and Sudheendra Vijayanarasimhan and Lora Aroyo and Vikas Yadav and Sarah Chakera and Ashwin Kakarla and Vilobh Meshram and Karol Gregor and Gabriela Botea and Evan Senter and Dawei Jia and Geza Kovacs and Neha Sharma and Sebastien Baur and Kai Kang and Yifan He and Lin Zhuo and Marija Kostelac and Itay Laish and Songyou Peng and Louis O'Bryan and Daniel Kasenberg and Girish Ramchandra Rao and Edouard Leurent and Biao Zhang and Sage Stevens and Ana Salazar and Ye Zhang and Ivan Lobov and Jake Walker and Allen Porter and Morgan Redshaw and Han Ke and Abhishek Rao and Alex Lee and Hoi Lam and Michael Moffitt and Jaeyoun Kim and Siyuan Qiao and Terry Koo and Robert Dadashi and Xinying Song and Mukund Sundararajan and Peng Xu and Chizu Kawamoto and Yan Zhong and Clara Barbu and Apoorv Reddy and Mauro Verzetti and Leon Li and George Papamakarios and Hanna Klimczak-Plucińska and Mary Cassin and Koray Kavukcuoglu and Rigel Swavely and Alain Vaucher and Jeffrey Zhao and Ross Hemsley and Michael Tschannen and Heming Ge and Gaurav Menghani and Yang Yu and Natalie Ha and Wei He and Xiao Wu and Maggie Song and Rachel Sterneck and Stefan Zinke and Dan A. Calian and Annie Marsden and Alejandro Cruzado Ruiz and Matteo Hessel and Almog Gueta and Benjamin Lee and Brian Farris and Manish Gupta and Yunjie Li and Mohammad Saleh and Vedant Misra and Kefan Xiao and Piermaria Mendolicchio and Gavin Buttimore and Varvara Krayvanova and Nigamaa Nayakanti and Matthew Wiethoff and Yash Pande and Azalia Mirhoseini and Ni Lao and Jasmine Liu and Yiqing Hua and Angie Chen and Yury Malkov and Dmitry Kalashnikov and Shubham Gupta and Kartik Audhkhasi and Yuexiang Zhai and Sudhindra Kopalle and Prateek Jain and Eran Ofek and Clemens Meyer and Khuslen Baatarsukh and Hana Strejček and Jun Qian and James Freedman and Ricardo Figueira and Michal Sokolik and Olivier Bachem and Raymond Lin and Dia Kharrat and Chris Hidey and Pingmei Xu and Dennis Duan and Yin Li and Muge Ersoy and Richard Everett and Kevin Cen and Rebeca Santamaria-Fernandez and Amir Taubenfeld and Ian Mackinnon and Linda Deng and Polina Zablotskaia and Shashank Viswanadha and Shivanker Goel and Damion Yates and Yunxiao Deng and Peter Choy and Mingqing Chen and Abhishek Sinha and Alex Mossin and Yiming Wang and Arthur Szlam and Susan Hao and Paul Kishan Rubenstein and Metin Toksoz-Exley and Miranda Aperghis and Yin Zhong and Junwhan Ahn and Michael Isard and Olivier Lacombe and Florian Luisier and Chrysovalantis Anastasiou and Yogesh Kalley and Utsav Prabhu and Emma Dunleavy and Shaan Bijwadia and Justin Mao-Jones and Kelly Chen and Rama Pasumarthi and Emily Wood and Adil Dostmohamed and Nate Hurley and Jiri Simsa and Alicia Parrish and Mantas Pajarskas and Matt Harvey and Ondrej Skopek and Yony Kochinski and Javier Rey and Verena Rieser and Denny Zhou and Sun Jae Lee and Trilok Acharya and Guowang Li and Joe Jiang and Xiaofan Zhang and Bryant Gipson and Ethan Mahintorabi and Marco Gelmi and Nima Khajehnouri and Angel Yeh and Kayi Lee and Loic Matthey and Leslie Baker and Trang Pham and Han Fu and Alex Pak and Prakhar Gupta and Cristina Vasconcelos and Adam Sadovsky and Brian Walker and Sissie Hsiao and Patrik Zochbauer and Andreea Marzoca and Noam Velan and Junhao Zeng and Gilles Baechler and Danny Driess and Divya Jain and Yanping Huang and Lizzie Tao and John Maggs and Nir Levine and Jon Schneider and Erika Gemzer and Samuel Petit and Shan Han and Zach Fisher and Dustin Zelle and Courtney Biles and Eugene Ie and Asya Fadeeva and Casper Liu and Juliana Vicente Franco and Adrian Collister and Hao Zhang and Renshen Wang and Ruizhe Zhao and Leandro Kieliger and Kurt Shuster and Rui Zhu and Boqing Gong and Lawrence Chan and Ruoxi Sun and Sujoy Basu and Roland Zimmermann and Jamie Hayes and Abhishek Bapna and Jasper Snoek and Weel Yang and Puranjay Datta and Jad Al Abdallah and Kevin Kilgour and Lu Li and SQ Mah and Yennie Jun and Morgane Rivière and Abhijit Karmarkar and Tammo Spalink and Tao Huang and Lucas Gonzalez and Duc-Hieu Tran and Averi Nowak and John Palowitch and Martin Chadwick and Ellie Talius and Harsh Mehta and Thibault Sellam and Philipp Fränken and Massimo Nicosia and Kyle He and Aditya Kini and David Amos and Sugato Basu and Harrison Jobe and Eleni Shaw and Qiantong Xu and Colin Evans and Daisuke Ikeda and Chaochao Yan and Larry Jin and Lun Wang and Sachin Yadav and Ilia Labzovsky and Ramesh Sampath and Ada Ma and Candice Schumann and Aditya Siddhant and Rohin Shah and John Youssef and Rishabh Agarwal and Natalie Dabney and Alessio Tonioni and Moran Ambar and Jing Li and Isabelle Guyon and Benny Li and David Soergel and Boya Fang and Georgi Karadzhov and Cristian Udrescu and Trieu Trinh and Vikas Raunak and Seb Noury and Dee Guo and Sonal Gupta and Mara Finkelstein and Denis Petek and Lihao Liang and Greg Billock and Pei Sun and David Wood and Yiwen Song and Xiaobin Yu and Tatiana Matejovicova and Regev Cohen and Kalyan Andra and David D'Ambrosio and Zhiwei Deng and Vincent Nallatamby and Ebrahim Songhori and Rumen Dangovski and Andrew Lampinen and Pankil Botadra and Adam Hillier and Jiawei Cao and Nagabhushan Baddi and Adhi Kuncoro and Toshihiro Yoshino and Ankit Bhagatwala and Marcáurelio Ranzato and Rylan Schaeffer and Tianlin Liu and Shuai Ye and Obaid Sarvana and John Nham and Chenkai Kuang and Isabel Gao and Jinoo Baek and Shubham Mittal and Ayzaan Wahid and Anita Gergely and Bin Ni and Josh Feldman and Carrie Muir and Pascal Lamblin and Wolfgang Macherey and Ethan Dyer and Logan Kilpatrick and Víctor Campos and Mukul Bhutani and Stanislav Fort and Yanif Ahmad and Aliaksei Severyn and Kleopatra Chatziprimou and Oleksandr Ferludin and Mason Dimarco and Aditya Kusupati and Joe Heyward and Dan Bahir and Kevin Villela and Katie Millican and Dror Marcus and Sanaz Bahargam and Caglar Unlu and Nicholas Roth and Zichuan Wei and Siddharth Gopal and Deepanway Ghoshal and Edward Lee and Sharon Lin and Jennie Lees and Dayeong Lee and Anahita Hosseini and Connie Fan and Seth Neel and Marcus Wu and Yasemin Altun and Honglong Cai and Enrique Piqueras and Josh Woodward and Alessandro Bissacco and Salem Haykal and Mahyar Bordbar and Prasha Sundaram and Sarah Hodkinson and Daniel Toyama and George Polovets and Austin Myers and Anu Sinha and Tomer Levinboim and Kashyap Krishnakumar and Rachita Chhaparia and Tatiana Sholokhova and Nitesh Bharadwaj Gundavarapu and Ganesh Jawahar and Haroon Qureshi and Jieru Hu and Nikola Momchev and Matthew Rahtz and Renjie Wu and Aishwarya P S and Kedar Dhamdhere and Meiqi Guo and Umang Gupta and Ali Eslami and Mariano Schain and Michiel Blokzijl and David Welling and Dave Orr and Levent Bolelli and Nicolas Perez-Nieves and Mikhail Sirotenko and Aman Prasad and Arjun Kar and Borja De Balle Pigem and Tayfun Terzi and Gellért Weisz and Dipankar Ghosh and Aditi Mavalankar and Dhruv Madeka and Kaspar Daugaard and Hartwig Adam and Viraj Shah and Dana Berman and Maggie Tran and Steven Baker and Ewa Andrejczuk and Grishma Chole and Ganna Raboshchuk and Mahdi Mirzazadeh and Thais Kagohara and Shimu Wu and Christian Schallhart and Bernett Orlando and Chen Wang and Alban Rrustemi and Hao Xiong and Hao Liu and Arpi Vezer and Nolan Ramsden and Shuo-yiin Chang and Sidharth Mudgal and Yan Li and Nino Vieillard and Yedid Hoshen and Farooq Ahmad and Ambrose Slone and Amy Hua and Natan Potikha and Mirko Rossini and Jon Stritar and Sushant Prakash and Zifeng Wang and Xuanyi Dong and Alireza Nazari and Efrat Nehoran and Kaan Tekelioglu and Yinxiao Li and Kartikeya Badola and Tom Funkhouser and Yuanzhen Li and Varun Yerram and Ramya Ganeshan and Daniel Formoso and Karol Langner and Tian Shi and Huijian Li and Yumeya Yamamori and Amayika Panda and Alaa Saade and Angelo Scorza Scarpati and Chris Breaux and CJ Carey and Zongwei Zhou and Cho-Jui Hsieh and Sophie Bridgers and Alena Butryna and Nishesh Gupta and Vaibhav Tulsyan and Sanghyun Woo and Evgenii Eltyshev and Will Grathwohl and Chanel Parks and Seth Benjamin and Rina Panigrahy and Shenil Dodhia and Daniel De Freitas and Chris Sauer and Will Song and Ferran Alet and Jackson Tolins and Cosmin Paduraru and Xingyi Zhou and Brian Albert and Zizhao Zhang and Lei Shu and Mudit Bansal and Sarah Nguyen and Amir Globerson and Owen Xiao and James Manyika and Tom Hennigan and Rong Rong and Josip Matak and Anton Bakalov and Ankur Sharma and Danila Sinopalnikov and Andrew Pierson and Stephen Roller and Geoff Brown and Mingcen Gao and Toshiyuki Fukuzawa and Amin Ghafouri and Kenny Vassigh and Iain Barr and Zhicheng Wang and Anna Korsun and Rajesh Jayaram and Lijie Ren and Tim Zaman and Samira Khan and Yana Lunts and Dan Deutsch and Dave Uthus and Nitzan Katz and Masha Samsikova and Amr Khalifa and Nikhil Sethi and Jiao Sun and Luming Tang and Uri Alon and Xianghong Luo and Dian Yu and Abhishek Nayyar and Bryce Petrini and Will Truong and Vincent Hellendoorn and Nikolai Chinaev and Chris Alberti and Wei Wang and Jingcao Hu and Vahab Mirrokni and Ananth Balashankar and Avia Aharon and Aahil Mehta and Ahmet Iscen and Joseph Kready and Lucas Manning and Anhad Mohananey and Yuankai Chen and Anshuman Tripathi and Allen Wu and Igor Petrovski and Dawsen Hwang and Martin Baeuml and Shreyas Chandrakaladharan and Yuan Liu and Rey Coaguila and Maxwell Chen and Sally Ma and Pouya Tafti and Susheel Tatineni and Terry Spitz and Jiayu Ye and Paul Vicol and Mihaela Rosca and Adrià Puigdomènech and Zohar Yahav and Sanjay Ghemawat and Hanzhao Lin and Phoebe Kirk and Zaid Nabulsi and Sergey Brin and Bernd Bohnet and Ken Caluwaerts and Aditya Srikanth Veerubhotla and Dan Zheng and Zihang Dai and Petre Petrov and Yichong Xu and Ramin Mehran and Zhuo Xu and Luisa Zintgraf and Jiho Choi and Spurthi Amba Hombaiah and Romal Thoppilan and Sashank Reddi and Lukasz Lew and Li Li and Kellie Webster and KP Sawhney and Lampros Lamprou and Siamak Shakeri and Mayank Lunayach and Jianmin Chen and Sumit Bagri and Alex Salcianu and Ying Chen and Yani Donchev and Charlotte Magister and Signe Nørly and Vitor Rodrigues and Tomas Izo and Hila Noga and Joe Zou and Thomas Köppe and Wenxuan Zhou and Kenton Lee and Xiangzhu Long and Danielle Eisenbud and Anthony Chen and Connor Schenck and Chi Ming To and Peilin Zhong and Emanuel Taropa and Minh Truong and Omer Levy and Danilo Martins and Zhiyuan Zhang and Christopher Semturs and Kelvin Zhang and Alex Yakubovich and Pol Moreno and Lara McConnaughey and Di Lu and Sam Redmond and Lotte Weerts and Yonatan Bitton and Tiziana Refice and Nicolas Lacasse and Arthur Conmy and Corentin Tallec and Julian Odell and Hannah Forbes-Pollard and Arkadiusz Socala and Jonathan Hoech and Pushmeet Kohli and Alanna Walton and Rui Wang and Mikita Sazanovich and Kexin Zhu and Andrei Kapishnikov and Rich Galt and Matthew Denton and Ben Murdoch and Caitlin Sikora and Kareem Mohamed and Wei Wei and Uri First and Tim McConnell and Luis C. Cobo and James Qin and Thi Avrahami and Daniel Balle and Yu Watanabe and Annie Louis and Adam Kraft and Setareh Ariafar and Yiming Gu and Eugénie Rives and Charles Yoon and Andrei Rusu and James Cobon-Kerr and Chris Hahn and Jiaming Luo and Yuvein and Zhu and Niharika Ahuja and Rodrigo Benenson and Raphaël Lopez Kaufman and Honglin Yu and Lloyd Hightower and Junlin Zhang and Darren Ni and Lisa Anne Hendricks and Gabby Wang and Gal Yona and Lalit Jain and Pablo Barrio and Surya Bhupatiraju and Siva Velusamy and Allan Dafoe and Sebastian Riedel and Tara Thomas and Zhe Yuan and Mathias Bellaiche and Sheena Panthaplackel and Klemen Kloboves and Sarthak Jauhari and Canfer Akbulut and Todor Davchev and Evgeny Gladchenko and David Madras and Aleksandr Chuklin and Tyrone Hill and Quan Yuan and Mukundan Madhavan and Luke Leonhard and Dylan Scandinaro and Qihang Chen and Ning Niu and Arthur Douillard and Bogdan Damoc and Yasumasa Onoe and Fabian Pedregosa and Fred Bertsch and Chas Leichner and Joseph Pagadora and Jonathan Malmaud and Sameera Ponda and Andy Twigg and Oleksii Duzhyi and Jingwei Shen and Miaosen Wang and Roopal Garg and Jing Chen and Utku Evci and Jonathan Lee and Leon Liu and Koji Kojima and Masa Yamaguchi and Arunkumar Rajendran and AJ Piergiovanni and Vinodh Kumar Rajendran and Marco Fornoni and Gabriel Ibagon and Harry Ragan and Sadh MNM Khan and John Blitzer and Andrew Bunner and Guan Sun and Takahiro Kosakai and Scott Lundberg and Ndidi Elue and Kelvin Guu and SK Park and Jane Park and Arunachalam Narayanaswamy and Chengda Wu and Jayaram Mudigonda and Trevor Cohn and Hairong Mu and Ravi Kumar and Laura Graesser and Yichi Zhang and Richard Killam and Vincent Zhuang and Mai Giménez and Wael Al Jishi and Ruy Ley-Wild and Alex Zhai and Kazuki Osawa and Diego Cedillo and Jialu Liu and Mayank Upadhyay and Marcin Sieniek and Roshan Sharma and Tom Paine and Anelia Angelova and Sravanti Addepalli and Carolina Parada and Kingshuk Majumder and Avery Lamp and Sanjiv Kumar and Xiang Deng and Artiom Myaskovsky and Tea Sabolić and Jeffrey Dudek and Sarah York and Félix de Chaumont Quitry and Jiazhong Nie and Dee Cattle and Alok Gunjan and Bilal Piot and Waleed Khawaja and Seojin Bang and Simon Wang and Siavash Khodadadeh and Raghavender R and Praynaa Rawlani and Richard Powell and Kevin Lee and Johannes Griesser and GS Oh and Cesar Magalhaes and Yujia Li and Simon Tokumine and Hadas Natalie Vogel and Dennis Hsu and Arturo BC and Disha Jindal and Matan Cohen and Zi Yang and Junwei Yuan and Dario de Cesare and Tony Bruguier and Jun Xu and Monica Roy and Alon Jacovi and Dan Belov and Rahul Arya and Phoenix Meadowlark and Shlomi Cohen-Ganor and Wenting Ye and Patrick Morris-Suzuki and Praseem Banzal and Gan Song and Pranavaraj Ponnuramu and Fred Zhang and George Scrivener and Salah Zaiem and Alif Raditya Rochman and Kehang Han and Badih Ghazi and Kate Lee and Shahar Drath and Daniel Suo and Antonious Girgis and Pradeep Shenoy and Duy Nguyen and Douglas Eck and Somit Gupta and Le Yan and Joao Carreira and Anmol Gulati and Ruoxin Sang and Daniil Mirylenka and Emma Cooney and Edward Chou and Mingyang Ling and Cindy Fan and Ben Coleman and Guilherme Tubone and Ravin Kumar and Jason Baldridge and Felix Hernandez-Campos and Angeliki Lazaridou and James Besley and Itay Yona and Neslihan Bulut and Quentin Wellens and AJ Pierigiovanni and Jasmine George and Richard Green and Pu Han and Connie Tao and Geoff Clark and Chong You and Abbas Abdolmaleki and Justin Fu and Tongzhou Chen and Ashwin Chaugule and Angad Chandorkar and Altaf Rahman and Will Thompson and Penporn Koanantakool and Mike Bernico and Jie Ren and Andrey Vlasov and Sergei Vassilvitskii and Maciej Kula and Yizhong Liang and Dahun Kim and Yangsibo Huang and Chengxi Ye and Dmitry Lepikhin and Wesley Helmholz},
      year={2025},
      eprint={2507.06261},
      archivePrefix={arXiv},
      primaryClass={cs.CL},
      url={https://arxiv.org/abs/2507.06261}, 
}

@inproceedings{agarwal_many-shot_2024,
	title = {Many-{Shot} {In}-{Context} {Learning}},
	url = {https://openreview.net/forum?id=AB6XpMzvqH},
	booktitle = {The {Thirty}-eighth {Annual} {Conference} on {Neural} {Information} {Processing} {Systems}},
	author = {Agarwal, Rishabh and Singh, Avi and Zhang, Lei M. and Bohnet, Bernd and Rosias, Luis and Chan, Stephanie C. Y. and Zhang, Biao and Anand, Ankesh and Abbas, Zaheer and Nova, Azade and Co-Reyes, John D. and Chu, Eric and Behbahani, Feryal and Faust, Aleksandra and Larochelle, Hugo},
	year = {2024},
}

@inproceedings{ibomnlp,
    title = "Ibom {NLP}: A Step Toward Inclusive Natural Language Processing for {N}igeria{'}s Minority Languages",
    author = "Kalejaiye, Oluwadara  and
      Beyene, Luel Hagos  and
      Adelani, David Ifeoluwa  and
      Edet, Mmekut-mfon Gabriel  and
      Akpan, Aniefon Daniel  and
      Urua, Eno-Abasi  and
      Andy, Anietie",
    editor = "Inui, Kentaro  and
      Sakti, Sakriani  and
      Wang, Haofen  and
      Wong, Derek F.  and
      Bhattacharyya, Pushpak  and
      Banerjee, Biplab  and
      Ekbal, Asif  and
      Chakraborty, Tanmoy  and
      Singh, Dhirendra Pratap",
    booktitle = "Proceedings of the 14th International Joint Conference on Natural Language Processing and the 4th Conference of the Asia-Pacific Chapter of the Association for Computational Linguistics",
    month = dec,
    year = "2025",
    address = "Mumbai, India",
    publisher = "The Asian Federation of Natural Language Processing and The Association for Computational Linguistics",
    url = "https://aclanthology.org/2025.ijcnlp-long.22/",
    pages = "372--382",
    ISBN = "979-8-89176-298-5"
}

@inproceedings{marashian2025,
    title = "From Priest to Doctor: Domain Adaptation for Low-Resource Neural Machine Translation",
    author = "Marashian, Ali  and
      Rice, Enora  and
      Gessler, Luke  and
      Palmer, Alexis  and
      von der Wense, Katharina",
    editor = "Rambow, Owen  and
      Wanner, Leo  and
      Apidianaki, Marianna  and
      Al-Khalifa, Hend  and
      Eugenio, Barbara Di  and
      Schockaert, Steven",
    booktitle = "Proceedings of the 31st International Conference on Computational Linguistics",
    month = jan,
    year = "2025",
    address = "Abu Dhabi, UAE",
    publisher = "Association for Computational Linguistics",
    url = "https://aclanthology.org/2025.coling-main.472/",
    pages = "7087--7098"
}

@article{nllb-24,
    author="{NLLB Team} and Costa-juss{\`a}, Marta R. and Cross, James and {\c{C}}elebi, Onur and Elbayad, Maha and Heafield, Kenneth and Heffernan, Kevin and Kalbassi, Elahe and Lam, Janice and Licht, Daniel and Maillard, Jean and Sun, Anna and Wang, Skyler and Wenzek, Guillaume and Youngblood, Al and Akula, Bapi and Barrault, Loic and Gonzalez, Gabriel Mejia and Hansanti, Prangthip and Hoffman, John and Jarrett, Semarley and Sadagopan, Kaushik Ram and Rowe, Dirk and Spruit, Shannon and Tran, Chau and Andrews, Pierre and Ayan, Necip Fazil and Bhosale, Shruti and Edunov, Sergey and Fan, Angela and Gao, Cynthia and Goswami, Vedanuj and Guzm{\'a}n, Francisco and Koehn, Philipp and Mourachko, Alexandre and Ropers, Christophe and Saleem, Safiyyah and Schwenk, Holger and Wang, Jeff",
    title="Scaling neural machine translation to 200 languages",
    journal="Nature",
    year="2024",
    volume="630",
    number="8018",
    pages="841--846",
    issn="1476-4687",
    doi="10.1038/s41586-024-07335-x",
    url="https://doi.org/10.1038/s41586-024-07335-x"
}

@inproceedings{pei_2025_understanding,
    title = "Understanding In-Context Machine Translation for Low-Resource Languages: A Case Study on {M}anchu",
    author = "Pei, Renhao  and
      Liu, Yihong  and
      Lin, Peiqin  and
      Yvon, Fran{\c{c}}ois  and
      Schuetze, Hinrich",
    editor = "Che, Wanxiang  and
      Nabende, Joyce  and
      Shutova, Ekaterina  and
      Pilehvar, Mohammad Taher",
    booktitle = "Proceedings of the 63rd Annual Meeting of the Association for Computational Linguistics (Volume 1: Long Papers)",
    month = jul,
    year = "2025",
    address = "Vienna, Austria",
    publisher = "Association for Computational Linguistics",
    url = "https://aclanthology.org/2025.acl-long.429/",
    doi = "10.18653/v1/2025.acl-long.429",
    pages = "8767--8788",
    ISBN = "979-8-89176-251-0"
}

@article{
luo2024incontext,
title={In-context Learning with Retrieved Demonstrations for Language Models: A Survey},
author={Man Luo and Xin Xu and Yue Liu and Panupong Pasupat and Mehran Kazemi},
journal={Transactions on Machine Learning Research},
issn={2835-8856},
year={2024},
url={https://openreview.net/forum?id=NQPo8ZhQPa},
note={Survey Certification}
}

@misc{ghazvininejad2023,
      title={Dictionary-based Phrase-level Prompting of Large Language Models for Machine Translation}, 
      author={Marjan Ghazvininejad and Hila Gonen and Luke Zettlemoyer},
      year={2023},
      eprint={2302.07856},
      archivePrefix={arXiv},
      primaryClass={cs.CL},
      url={https://arxiv.org/abs/2302.07856}, 
}

@inproceedings{adelani-etal-2022-thousand,
    title = "A Few Thousand Translations Go a Long Way! Leveraging Pre-trained Models for {A}frican News Translation",
    author = "Adelani, David Ifeoluwa  and
      Alabi, Jesujoba Oluwadara  and
      Fan, Angela  and
      Kreutzer, Julia  and
      Shen, Xiaoyu  and
      Reid, Machel  and
      Ruiter, Dana  and
      Klakow, Dietrich  and
      Nabende, Peter  and
      Chang, Ernie  and
      Gwadabe, Tajuddeen  and
      Sackey, Freshia  and
      Dossou, Bonaventure F. P.  and
      Emezue, Chris  and
      Leong, Colin  and
      Beukman, Michael  and
      Muhammad, Shamsuddeen H.  and
      Jarso, Guyo D.  and
      Yousuf, Oreen  and
      Niyongabo Rubungo, Andre N.  and
      Hacheme, Gilles  and
      Wairagala, Eric Peter  and
      Nasir, Muhammad Umair  and
      Ajibade, Benjamin A.  and
      Ajayi, Tunde Oluwaseyi  and
      Gitau, Yvonne Wambui  and
      Abbott, Jade  and
      Ahmed, Mohamed  and
      Ochieng, Millicent  and
      Aremu, Anuoluwapo  and
      Ogayo, Perez  and
      Mukiibi, Jonathan  and
      Ouoba Kabore, Fatoumata  and
      Kalipe, Godson Koffi  and
      Mbaye, Derguene  and
      Tapo, Allahsera Auguste  and
      Memdjokam Koagne, Victoire M.  and
      Munkoh-Buabeng, Edwin  and
      Wagner, Valencia  and
      Abdulmumin, Idris  and
      Awokoya, Ayodele  and
      Buzaaba, Happy  and
      Sibanda, Blessing  and
      Bukula, Andiswa  and
      Manthalu, Sam",
    editor = "Carpuat, Marine  and
      de Marneffe, Marie-Catherine  and
      Meza Ruiz, Ivan Vladimir",
    booktitle = "Proceedings of the 2022 Conference of the North American Chapter of the Association for Computational Linguistics: Human Language Technologies",
    month = jul,
    year = "2022",
    address = "Seattle, United States",
    publisher = "Association for Computational Linguistics",
    url = "https://aclanthology.org/2022.naacl-main.223/",
    doi = "10.18653/v1/2022.naacl-main.223",
    pages = "3053--3070"
}

@inproceedings{
tanzer2024a,
title={A Benchmark for Learning to Translate a New Language from One Grammar Book},
author={Garrett Tanzer and Mirac Suzgun and Eline Visser and Dan Jurafsky and Luke Melas-Kyriazi},
booktitle={The Twelfth International Conference on Learning Representations},
year={2024},
url={https://openreview.net/forum?id=tbVWug9f2h}
}

@inproceedings{zhang2023prompting,
  title={Prompting large language model for machine translation: A case study},
  author={Zhang, Biao and Haddow, Barry and Birch, Alexandra},
  booktitle={International conference on machine learning},
  pages={41092--41110},
  year={2023},
  organization={PMLR}
}

%\newpage
% ================================================================
% APPENDIX
% ================================================================
\clearpage
\appendix   

\section{Languages}
\label{app:languages}

We provide the details of the studied languages in Table~\ref{tab:languages}.

ICL examples are sampled from the devtest split (1,012 sentences), and evaluation is performed on the dev split (997 sentences).

For the Bible data, we use versions obtained from bible.com.
These texts are used strictly for research purposes without redistribution, in accordance with their terms of use.

\section{Prompt Templates}
\label{app:prompts}

All experiments use the following prompt format without any dictionary augmentation.

\paragraph{Many-shot prompt ($k \geq 1$ examples).}

\begin{small}
\begin{verbatim}
You are an expert translator. I am going to
give you one or more example pairs of text
snippets where the first is in {SRC} and the
second is a translation of the first snippet
into {TGT}. The sentences will be written
{SRC}: {example_src_1}
{TGT}: {example_tgt_1}
{SRC}: {example_src_2}
{TGT}: {example_tgt_2}
...
{SRC}: {example_src_k}
{TGT}: {example_tgt_k}
After the example pairs, I am going to provide
another sentence in {SRC} and I want you to
translate it into {TGT}. Give only the
translation, and no extra commentary,
formatting, or chattiness. Translate the text
from {SRC} to {TGT}.
{SRC}: {query}
{TGT}:
\end{verbatim}
\end{small}

\paragraph{Zero-shot prompt.}

\begin{small}
\begin{verbatim}
You are an expert translator. Translate the
text from {SRC} to {TGT}. Give only the
translation, and no extra commentary,
formatting, or chattiness.
{SRC}: {query}
{TGT}:
\end{verbatim}
\end{small}

\noindent
Here \texttt{\{SRC\}} and \texttt{\{TGT\}} are replaced with the full language names (e.g., ``English'', ``Ibibio'').
In the many-shot setting, the $k$ example pairs are drawn from the example pool, either by random sampling or BM25 retrieval (\autoref{app:retrieval}).

\section{Related Work}
\label{app:relatedwork}

While ICL~\citep{brown2020language} has been shown to be very effective on tasks such as machine translation~\citep{dong2024survey}, performance is highly sensitive to the choice of demonstrations \cite{luo2024incontext}, including factors such as example order~\cite{lu-etal-2022-fantastically}, diversity~\cite{li-qiu-2023-finding}, and difficulty~\cite{drozdov-etal-2023-parade}.  
In MT, \citet{agrawal-etal-2023-context} show that relevance-based retrieval improves few-shot ICL for high-resource languages.  
However, it remains unclear how example selection interacts with scale in the many-shot setting, particularly for extremely LRLs.

\citet{agarwal_many-shot_2024} challenge this paradigm by scaling ICL to hundreds or thousands of examples with Gemini~1.5~Pro, showing substantial improvements across summarization, reasoning, and MT, including low-resource translation into Bemba and Kurdish. \citet{salim2026beyond} investigate scaling in-context token budget to 1M tokens, but the evaluation is limited to smaller models like \texttt{Qwen 2.5 7B} with worse performance for larger shots. 
Our work extends the many-shot setting to a systematic study of 10 extremely LRLs, with a particular focus on retrieval-based example selection and domain effects.

\section{Retrieval Details}
\label{app:retrieval}

BM25 example selection uses BM25~\cite{10.1561/1500000019} with default parameters. Indexing and querying both use the English source side; English sentences are tokenized by lowercasing, replacing punctuation with whitespace, removing digits, and splitting on whitespace. For each test sentence, we score every pool sentence with BM25 and select the top-$k$ by descending score; the corresponding parallel pairs are then placed in the prompt according to one of three orderings:

\begin{itemize}\setlength{\itemsep}{0pt}
\item \textbf{Default} (used in the main BM25 results, \S\ref{sec:selection}): randomly shuffle the top-$k$ pairs with a fixed seed.
\item \textbf{Similar first} (Table~\ref{tab:ordering_retrieval_stacked}): place pairs in descending BM25 score, so the most-similar pair appears at the start of the prompt (farthest from the query).
\item \textbf{Dissimilar first} (Table~\ref{tab:ordering_retrieval_stacked}): reverse of \textit{similar first}, so the most-similar pair appears immediately before the query.
\end{itemize}
% (see \S\ref{sec:protocol}).

\section{Full Model Results}
\label{app:full_results}

\paragraph{eng$\rightarrow$X.}
Each table reports Random (in-domain), BM25 retrieval, BM25$-$Random delta ($\Delta$), and Bible (cross-domain) results across all shot counts.
Gemini~2.5~Flash: \autoref{tab:bible_vs_indomain_gemini_chrfpp} (chrF++) and \autoref{tab:bible_vs_indomain_gemini_spbleu} (spBLEU).
GPT-4.1: \autoref{tab:bible_vs_indomain_gpt4_chrfpp} (chrF++) and \autoref{tab:bible_vs_indomain_gpt4_spbleu} (spBLEU).
Llama~3.3~70B: \autoref{tab:bible_vs_indomain_llama_chrfpp} (chrF++) and \autoref{tab:bible_vs_indomain_llama_spbleu} (spBLEU).
Gemma~3~27B: \autoref{tab:bible_vs_indomain_gemma3_27b_chrfpp} (chrF++) and \autoref{tab:bible_vs_indomain_gemma3_27b_spbleu} (spBLEU).
All eight tables share the following formatting: the six languages with a Bible translation appear on the left; the four without (\xmark) on the right. \textbf{Bold} marks the best score per column. \textcolor{gray}{Gray} indicates the 0-shot baseline. $\Delta$ values are \textcolor{green!60!black}{green} when BM25 outperforms Random and \textcolor{red!70!black}{red} otherwise. \textcolor{blue!70!black}{Blue} at the bottom shows the gain from Random at 1,000 shots over 0-shot.

\paragraph{X$\rightarrow$eng (random selection only).}
\autoref{tab:x2eng_stacked_chrfpp} (chrF++) and \autoref{tab:x2eng_stacked_spbleu} (spBLEU) report X$\rightarrow$eng scaling results with random selection across all four models. \textcolor{gray}{Gray} indicates the 0-shot baseline; \textbf{bold} marks the best per column. The \textit{gain} row shows the ratio of the best score to 0-shot (\textcolor{blue!70!black}{blue}; \textbf{\textcolor{green!60!black}{green bold}} for ${\geq}2\times$).

Missing entries (--) indicate runs that exceeded the model's context window.
% or failed quality checks.

\section{Ordering Comparison}
\label{app:ordering}

We test whether the order in which examples appear in the prompt affects translation quality. Table~\ref{tab:ordering_retrieval_stacked} varies the ordering of retrieved examples (BM25 and dense retrieval); Table~\ref{tab:ordering_length_stacked} sorts randomly selected examples by sentence length. In both cases, differences are negligible and no strategy consistently wins.

\section{Fine-tuning Comparison}
\label{app:lora}

\autoref{tab:lora_full} reports per-shot, per-direction spBLEU and chrF++ for the case study in \S\ref{sec:lora}, in the same stacked format as \autoref{tab:ordering_length_stacked}. \texttt{Gemini 2.5 Flash} is included as a closed-API reference; its BM25 values are copied from \autoref{tab:bible_vs_indomain_gemini_chrfpp} and \autoref{tab:bible_vs_indomain_gemini_spbleu}. Aggregate averages reported in the main text are computed over the two directions.

\paragraph{Cost considerations.}
Translating the 997-sentence query set using 250 shots ICL costs approximately \$2--\$3 per direction with \texttt{Gemini 2.5 Flash} (Batch API rate) for typical Latin-script LRLs (e.g., Mauritian Creole, Ladin, Efik, Ibibio), and up to \$12 for under-represented scripts such as Tifinagh; \texttt{GPT-4.1} costs approximately \$46 per direction at standard pricing on \mbox{eng$\rightarrow$Efik}, or \$23 at the Azure Batch rate we used in our experiments.

The cost of fine-tuning an open-weight alternative depends on the choice of model and compute budget. Renting a single A100 80GB GPU typically costs more than \$2 per hour by most providers such as Lambda (A100 cost \$2.79) , but the total cost scales with model size and training-pass duration. Deployment cost for self-hosted open-weight models is harder to characterize, as it depends on the user's available hardware and throughput requirements; matching the translation quality of the strongest closed-API models in our experiments would typically require open-weight models substantially larger than those we fine-tuned, which are themselves nontrivial to self-host efficiently.

For practitioners with access to a closed-API model, it delivers both the strongest quality we measure (e.g., 35.4 vs.\ 30.1 chrF++ at 250 shots on \mbox{eng$\rightarrow$Efik}; \autoref{tab:lora_full}) and a cost on the order of a few dollars per test set, accessible to low-resource language communities without infrastructure investment. For open-weight-only deployments, based on our experimental results, you should only finetune without using ICL if you need to reuse the model extensively and only require a relatively mediocre translation performance. If you are only translating a small amount, then ICL alone is sufficient to achieve the same performance.(Based on our experiments, open-source models only need 10-25 shots to achieve the performance of model fine-tuned using full 1012 sentence example pools, which is much cheaper than the 250 shots we discussed.). If you want to have the best possible performance, you still need to add ICL, and in this case, fine-tuning may no longer be necessary, because in our experiments, under 250-shot ICL, the performance with and without loRA fine-tuning was very similar.

The total of more than \$30{,}000 reported in our Limitations corresponds to our actual API billing across the full experimental program, including pilot and exploration runs beyond the configurations reported in the main results. Because input prompt length---and thus per-token API cost---scales with the number of shots, less than 25\% of this total was spent on experiments at $k \leq 250$; the remaining \mbox{${\sim}$75\%} comes from high-shot scaling experiments at $k \in \{500, 750, 1000\}$ (only the $k = 500$ and $k = 1000$ endpoints are reported in the main scaling figures).

\section{Dictionary Augmentation}
\label{app:dict}

Bilingual dictionaries are a common augmentation for prompt-based MT in low-resource settings, with prior work reporting gains at low shot counts~\cite{ghazvininejad2023, zhang_teaching_2024, pei_2025_understanding}. These gains, however, depend on the availability of comprehensive, high-quality dictionaries --- a condition that is often not met for extremely low-resource languages, whose digital dictionaries (where they exist at all) tend to be small, incomplete, or prone to OCR and typographical errors. We test whether dictionary augmentation persists in the many-shot regime under such realistic LRL conditions on \mbox{eng$\rightarrow$Efik} and \mbox{eng$\rightarrow$Anaang}, the two directions for which we have a digital dictionary, comparing a 25-shot BM25 prompt augmented with the dictionary against both the matched 25-shot baseline and a 50-shot baseline without the dictionary (\autoref{tab:dict_ablation}). At a matched 25-shot budget, the dictionary yields only marginal gains on \texttt{GPT-4.1} ($+0.55$ and $+0.05$ chrF++) and slight losses on \texttt{Gemini 2.5 Flash} ($-0.41$ and $-0.19$). At a matched prompt budget, adding the dictionary on top of 25 shots is consistently outperformed by simply scaling to 50 retrieved examples (all four $\Delta_{50}$ values negative). Two factors plausibly contribute: the limited coverage and quality of the digital dictionaries available for these languages, and the diminishing marginal value of a separate dictionary once retrieved examples already supply relevant target-language vocabulary. We therefore do not interpret these results as a refutation of dictionary augmentation in general, which remains useful at lower shot counts and with higher-coverage dictionaries.

\section{Qualitative Analysis}
\label{app:qualitative}

\paragraph{Per-language scaling patterns.}
We provide a post-hoc analysis of the language-specific scaling patterns observed in \S\ref{sec:scaling}. Following \citet{zebaze-etal-2025-context}, we view ICL gains as reflecting two largely independent bottlenecks: \emph{task and target-language anchoring}, which may be resolved with very few examples; and \emph{target-language generation quality}, which may continue to benefit from additional lexical, orthographic, and syntactic evidence. Under this interpretation, Tamazight's large 0$\rightarrow$1-shot gain is consistent with a script-anchoring effect: Tifinagh is severely under-represented in pretraining, so the model fails to produce target-script output at 0-shot (near-zero chrF++), and a single in-context example may suffice to anchor it to the correct target script. We verified this by inspecting actual outputs: at 0-shot, the model produces valid Tamazight content transliterated in Latin script rather than written in Tifinagh; a single Tifinagh in-context example switches the output to the correct script (Table~\ref{tab:tamazight_anchoring}). By contrast, Oro uses an extended Latin script (with diacritics) but remains extremely low-resource; its continued gains with more shots suggest that additional examples may provide useful target-language evidence beyond task identification. The remaining eight languages show intermediate scaling behavior (see \autoref{tab:bible_vs_indomain_gemini_chrfpp} for the full scaling curves). We emphasize that these are hypotheses supported by observed scaling patterns and language metadata, not causal claims.

\begin{table}[t]
\centering
\small
\begin{tabular}{llrr}
\toprule
\textbf{Language} & \textbf{Script} & \textbf{0$\rightarrow$1} & \textbf{1$\rightarrow$1000} \\
\midrule
Tamazight        & Tifinagh           & $+24.6$ & $+11.0$ \\
Oro              & Latin + diacritics & $+4.1$  & $+25.2$ \\
\bottomrule
\end{tabular}
\caption{0$\rightarrow$1 and 1$\rightarrow$1000 chrF++ gains for Tamazight and Oro (Random selection), computed from \autoref{tab:bible_vs_indomain_gemini_chrfpp}.}
\label{tab:per_lang_interpretation}
\end{table}

\begin{table}[t]
\centering
\small
\begin{tabular}{l p{0.24\textwidth}}
\toprule
\textbf{English (source)}      & The pilot was identified as Squadron Leader Dilokrit Pattavee. \\
\textbf{Reference (Tifinagh)}  & {\tifinaghfont ⵓⴼⴰⵏ ⵏⵏ ⴰⵎⵙⵙⵉⵢⵍ ⵉⴳⴰ ⴰⵎⵇⵇⵔⴰⵏ ⵏ ⵓⴷⴼⵔⵉⵔ ⴷⵉⵍⵓⴽⵔⵉⵜ ⴱⴰⵜⴰⴼⵉ.} \\
\textbf{\texttt{Gemini Flash} 0-shot} & Amhar ittwasn d Amghar n Tagrawt Dilokrit Pattavee. \\
\textbf{\texttt{Gemini Flash} 1-shot} & {\tifinaghfont ⴰⵏⴰⵢ ⵉⵜⵜⵡⴰⵙⵙⵏ ⴷ ⴰⵏⵙⵙⵉⵅⴼ ⵏ ⵜⴰⴳⵔⵓⵎⵜ ⴷⵉⵍⵓⴽⵔⵉⵜ ⴱⴰⵜⵜⴰⴼⵉ.} \\
\bottomrule
\end{tabular}
\caption{Illustrative example of script anchoring on \mbox{eng$\rightarrow$Tamazight}. At 0-shot, \texttt{Gemini 2.5 Flash} produces valid Tamazight content transliterated in Latin script; a single Tifinagh in-context example switches the output to Tifinagh. The proper name ``Dilokrit Pattavee'' is preserved across all three versions (left in Latin at 0-shot, transliterated to Tifinagh at 1-shot and in the reference), confirming the failure mode is target-script anchoring rather than translation quality.}
\label{tab:tamazight_anchoring}
\end{table}

\paragraph{Native-speaker assessment for Oro.}
To complement the aggregate chrF++ trends in \S\ref{sec:scaling}, we asked a native speaker of Oro to rate \mbox{eng$\rightarrow$Oro} translations from \texttt{Gemini 2.5 Flash} using Direct Assessment (DA, 0--100; higher is better) for a single FLORES+ sentence across shot counts (Table~\ref{tab:qualitative_oro}). The example illustrates that low-shot chrF++ gains can be misleading: at $k{\leq}10$ the model produces fluent-looking output that mimics the target-language orthography but is rated as not useful (DA${=}0$); only at $k{\geq}100$ does the translation become adequate to the native speaker. We observed a similar pattern for Ibibio. We report this as an illustrative example rather than a systematic human evaluation.

\begin{table*}[t]
\centering
\footnotesize
\setlength{\tabcolsep}{4pt}
\renewcommand{\arraystretch}{1.2}
\begin{tabular}{@{}l p{0.82\textwidth} c@{}}
\toprule
\textbf{Setting} & \textbf{Translation output} & \textbf{DA} \\
\midrule
\multicolumn{3}{@{}p{0.97\textwidth}@{}}{\textbf{Source (eng):} The valves opening allowed a pressure release for the system and oil flowed on a pad to a tank that can hold 55,000 barrels (2.3 million gallons).}\\
\addlinespace[2pt]
\multicolumn{3}{@{}p{0.97\textwidth}@{}}{\textbf{Reference (oro):} Ku inuẹ onu ku ovim ẹtei osi ọwọñọ kini ara ẹwẹ aga ọdọñọ ku ata ogbono uga omu okeme idọñọ izim miniọn otu ara 55,000 (miniọn gallon 2.3).}\\
\midrule
1-shot   & Nmeghe mkpọkọ ahụ mere ka nrụgide tọhapụ maka usoro ahụ, ma mmanụ wee reree n'elu ebe gaa na tankị nke nwere ike ijide barel puku iri ise na ise (galọn nde abụọ na puku narị atọ). & 0 \\
\addlinespace[2pt]
10-shot  & Ima valves egegin ọwọ ku ufene ifrẹ ọnọ imu udip ma ọkpọ-ọkpọ ọgọ ku pad ku tank omu ekeme imuọ 55,000 barrels (2.3 igbẹghẹ gallons). & 0 \\
\addlinespace[2pt]
100-shot & Kini ima valve egebuana, emu egeyak ulu ozighi ọnọ ubonokpọ onu, ma ara ọgọwọñ ku ufene edi ufene-mmọñ omu okeme imuọ mkpọ su barrel 55,000 (miniọn 2.3 galọn). & 50 \\
\addlinespace[2pt]
250-shot & Ima valve omu egedi ntuñọ ẹgeyak ntuñọ ulagha ọnọ ubonokpọ ma ara ọgọwẹ ku ogbogro aga udọñ omu okeme imuọ 55,000 barrel (miniọn 2.3 galọn). & 60 \\
\bottomrule
\end{tabular}
\caption{Illustrative native-speaker Direct Assessment (DA, 0--100; higher is better) of \mbox{eng$\rightarrow$Oro} translations from \texttt{Gemini 2.5 Flash} for a single FLORES+ sentence, across shot counts $k$. Low-shot outputs mimic the target-language orthography but are inadequate (DA${=}0$); adequacy emerges only at $k{\geq}100$.}
\label{tab:qualitative_oro}
\end{table*}

\begin{table*}[t]
\centering
\footnotesize
\setlength\tabcolsep{2.8pt}
\begin{threeparttable}
\begin{tabular}{lllllc}
\toprule
\textbf{Language} & \textbf{Family} & \textbf{Morphology} & \textbf{Script} & \textbf{Dataset} & \textbf{Has a Bible?} \\
\midrule
Emakhuwa (vmw)       & Atlantic–Congo / Bantu       & Agglutinative  & Latin    &  FLORES+\tnote{a}  & \cmark\\
Moroccan Tamazight (zgh) & Afro-Asiatic / Berber   & Fusional       & Tifinagh &  FLORES+\tnote{b}  & \xmark \\
Ladin (lld)          & Indo-European / Romance     & Fusional       & Latin    &  FLORES+\tnote{c}  & \xmark  \\
Mauritian Creole (mfe) & French Creole             & Analytic       & Latin    &  FLORES+\tnote{d}  & \cmark \\
Ay.\ Quechua (quy)  & Quechuan                    & Agglutinative  & Latin    & FLORES+\tnote{e}  & \cmark  \\
Sudanese Arabic (apd) & Afro-Asiatic / Semitic       & Fusional       & Arabic   & Sudanese-Flores\tnote{f} & \cmark \\
\midrule
Anaang (anw)         & Atlantic–Congo / Cross River  & Agglutinative  & Latin    & Ibom-NLP\tnote{g} & \xmark\\
Efik (efi)           & Atlantic–Congo / Cross River  & Agglutinative  & Latin    & Ibom-NLP\tnote{g} & \cmark \\
Ibibio (ibb)         & Atlantic–Congo / Cross River  & Agglutinative  & Latin    & Ibom-NLP\tnote{g} & \cmark\\
Oro (oro)            & Atlantic–Congo / Cross River  & Agglutinative  & Latin    & Ibom-NLP\tnote{g} & \cmark \\
\bottomrule
\end{tabular}
\begin{tablenotes}[para]
\footnotesize
\item[a] \cite{ali-etal-2024-expanding} 
\item[b] \cite{oktem-etal-2025-correcting}
\item[c] \cite{frontull-etal-2025-bringing}
\item[d] \cite{rajcoomar-2025-kozkreolmru}
\item[e] \cite{nllb-24}
\item[f] \cite{samil2026sudaneseflores}
\item[g] \cite{ibomnlp}
\end{tablenotes}
\end{threeparttable}
\caption{Overview of 10 target languages. The bottom four are from the Ibom region of Nigeria.}
\label{tab:languages}
\end{table*}

\begin{table}[ht]
\centering
\small
\setlength{\tabcolsep}{6pt}
\begin{tabular}{cl|cc}
\toprule
\textbf{Shots} & \textbf{Variant} & \multicolumn{2}{c}{\textbf{spBLEU\,/\,chrF++}} \\
\cmidrule(lr){3-4}
 & & efi & ibb \\
\midrule
\multirow{4}{*}{0}
  & Base \texttt{Gemma}        & 3.6/19.8 & 3.8/18.5 \\
  & + \textsc{LoRA(250)}       & 6.4/24.1 & 6.5/24.2 \\
  & + \textsc{LoRA(1012)}      & 8.4/27.3 & \textbf{8.0/26.4} \\
  & \texttt{Gemini 2.5 Flash}  & \textbf{9.1/31.3} & 6.3/25.0 \\
\cmidrule(lr){2-4}
\multirow{4}{*}{1}
  & Base \texttt{Gemma}        & 4.8/22.1 & 4.9/20.2 \\
  & + \textsc{LoRA(250)}       & 6.3/24.4 & 6.4/24.0 \\
  & + \textsc{LoRA(1012)}      & 8.9/27.8 & \textbf{8.7/26.9} \\
  & \texttt{Gemini 2.5 Flash}  & \textbf{10.1/32.1} & 7.5/26.0 \\
\cmidrule(lr){2-4}
\multirow{4}{*}{5}
  & Base \texttt{Gemma}        & 6.8/25.5 & 6.7/23.2 \\
  & + \textsc{LoRA(250)}       & 7.5/26.3 & 7.5/25.2 \\
  & + \textsc{LoRA(1012)}      & 9.3/28.5 & 9.1/27.5 \\
  & \texttt{Gemini 2.5 Flash}  & \textbf{11.5/33.5} & \textbf{10.1/28.7} \\
\cmidrule(lr){2-4}
\multirow{4}{*}{10}
  & Base \texttt{Gemma}        & 7.7/27.0 & 7.7/24.7 \\
  & + \textsc{LoRA(250)}       & 8.0/27.0 & 8.0/25.8 \\
  & + \textsc{LoRA(1012)}      & 9.9/28.9 & 9.5/27.7 \\
  & \texttt{Gemini 2.5 Flash}  & \textbf{11.9/33.7} & \textbf{11.5/30.1} \\
\cmidrule(lr){2-4}
\multirow{4}{*}{25}
  & Base \texttt{Gemma}        & 8.9/28.4 & 9.1/26.5 \\
  & + \textsc{LoRA(250)}       & 9.1/28.2 & 8.8/26.4 \\
  & + \textsc{LoRA(1012)}      & 10.3/29.3 & 9.9/28.1 \\
  & \texttt{Gemini 2.5 Flash}  & \textbf{12.6/34.4} & \textbf{13.1/31.7} \\
\cmidrule(lr){2-4}
\multirow{4}{*}{50}
  & Base \texttt{Gemma}        & 9.7/29.2 & 9.7/27.1 \\
  & + \textsc{LoRA(250)}       & 9.5/28.9 & 9.4/27.0 \\
  & + \textsc{LoRA(1012)}      & 10.7/29.7 & 10.4/28.6 \\
  & \texttt{Gemini 2.5 Flash}  & \textbf{13.5/35.2} & \textbf{13.9/32.8} \\
\cmidrule(lr){2-4}
\multirow{4}{*}{100}
  & Base \texttt{Gemma}        & 10.3/29.5 & 9.7/27.6 \\
  & + \textsc{LoRA(250)}       & 9.9/29.2 & 9.5/27.1 \\
  & + \textsc{LoRA(1012)}      & 10.7/29.6 & 10.6/28.8 \\
  & \texttt{Gemini 2.5 Flash}  & \textbf{13.4/35.1} & \textbf{14.7/33.7} \\
\cmidrule(lr){2-4}
\multirow{4}{*}{250}
  & Base \texttt{Gemma}        & 10.3/29.3 & 9.8/27.3 \\
  & + \textsc{LoRA(250)}       & 10.0/29.4 & 10.0/27.5 \\
  & + \textsc{LoRA(1012)}      & 10.9/30.1 & 10.4/28.5 \\
  & \texttt{Gemini 2.5 Flash}  & \textbf{13.8/35.4} & \textbf{15.2/34.2} \\
\bottomrule
\end{tabular}
\caption{Each cell reports \emph{spBLEU\,/\,chrF++} on \mbox{eng$\rightarrow$\{Efik, Ibibio\}} comparing \texttt{Gemma 3 27B} fine-tuning variants with \texttt{Gemini 2.5 Flash}, at 0 to 250 BM25-retrieved shots. \textbf{Bold} marks the variant with the highest score in that cell (spBLEU and chrF++ agree on the winner everywhere). \texttt{Gemini 2.5 Flash} cannot be LoRA-tuned through its API and is shown only as a closed-API reference at matched in-context counts.}
\label{tab:lora_full}
\end{table}

\begin{table*}[ht]
\centering
\footnotesize
\setlength{\tabcolsep}{6pt}
\begin{tabular}{ll|rrr|rr}
\toprule
\textbf{Model} & \textbf{Direction} & \textbf{25+dict} & \textbf{25} & \textbf{$\Delta_{25}$} & \textbf{50} & \textbf{$\Delta_{50}$} \\
\midrule
\multirow{2}{*}{\texttt{GPT-4.1}}
  & eng$\rightarrow$Efik   & 29.68 & 29.13 & $+0.55$ & 29.90 & $-0.22$ \\
  & eng$\rightarrow$Anaang & 25.63 & 25.58 & $+0.05$ & 26.12 & $-0.49$ \\
\multirow{2}{*}{\texttt{Gemini 2.5 Flash}}
  & eng$\rightarrow$Efik   & 33.94 & 34.35 & $-0.41$ & 35.23 & $-1.29$ \\
  & eng$\rightarrow$Anaang & 27.43 & 27.62 & $-0.19$ & 28.21 & $-0.78$ \\
\bottomrule
\end{tabular}
\caption{Dictionary augmentation versus scaling in-context examples (chrF++). ``25+dict'' prepends a bilingual word list to a 25-shot BM25 prompt; ``25'' and ``50'' are no-dict BM25 baselines. $\Delta_{25}$ reports (25+dict) $-$ (25): dictionary gain at matched shot count. $\Delta_{50}$ reports (25+dict) $-$ (50): dictionary gain versus reallocating the same prompt budget to more examples. $\Delta_{50}$ is negative across all four model--language pairs: at a matched prompt budget, adding a dictionary on top of 25 shots is consistently worse than simply scaling to 50 shots.}
\label{tab:dict_ablation}
\end{table*}

% --- T: Main Scaling Results (Gemini only, fine-grained) ---
\begin{table*}[t]
\centering
\footnotesize

\begin{tabular}{cl|cccccc|cccc}
\toprule
\textbf{Shots} & \textbf{Src} & \multicolumn{6}{c|}{chrF++ (Bible langs)} & \multicolumn{4}{c}{chrF++ (no Bible)} \\
\cmidrule(lr){3-8} \cmidrule(lr){9-12}
 & & efi & ibb & mfe & oro & quy & vmw & anw & lld & zgh & apd \\
\midrule
\multicolumn{2}{c|}{\textcolor{gray}{\textit{0-shot}}} & \textcolor{gray}{31.3} & \textcolor{gray}{25.0} & \textcolor{gray}{55.4} & \textcolor{gray}{13.6} & \textcolor{gray}{33.4} & \textcolor{gray}{26.8} & \textcolor{gray}{15.9} & \textcolor{gray}{38.9} & \textcolor{gray}{2.6} & \textcolor{gray}{43.9} \\
\midrule
\multirow{4}{*}{1}
  & Random  & 31.7 & 25.3 & 56.8 & 17.7 & 34.1 & 27.5 & 21.5 & 39.4 & 27.2 & \textbf{44.8} \\
  & BM25    & \textbf{32.1} & \textbf{26.0} & \textbf{57.5} & \textbf{20.2} & \textbf{34.5} & \textbf{28.7} & \textbf{22.7} & \textbf{43.8} & \textbf{28.6} & 44.6 \\
  & BM25$-$Random ($\triangle$) & \textcolor{green!60!black}{+0.3} & \textcolor{green!60!black}{+0.7} & \textcolor{green!60!black}{+0.7} & \textcolor{green!60!black}{+2.5} & \textcolor{green!60!black}{+0.3} & \textcolor{green!60!black}{+1.2} & \textcolor{green!60!black}{+1.2} & \textcolor{green!60!black}{+4.4} & \textcolor{green!60!black}{+1.4} & \textcolor{red!70!black}{$-$0.2} \\
  & Bible   & \textbf{32.2} & 25.6 & 55.9 & 16.2 & 33.6 & 27.7 & \xmark & \xmark & \xmark & \xmark \\
\midrule
\multirow{4}{*}{5}
  & Random  & 31.9 & 26.4 & 58.3 & 23.3 & 34.5 & 28.2 & 22.9 & 45.1 & 28.7 & 44.3 \\
\bottomrule
\end{tabular}

\caption{eng$\rightarrow$X results with Gemini 2.5 Flash (chrF++). Formatting conventions as described in \S\ref{app:full_results}.}
\label{tab:bible_vs_indomain_gemini_chrfpp}

\end{table*}

\begin{table*}[t]
\centering
\footnotesize

\begin{tabular}{cl|cccccc|cccc}
\toprule
\textbf{Shots} & \textbf{Src} & \multicolumn{6}{c|}{spBLEU (Bible langs)} & \multicolumn{4}{c}{spBLEU (no Bible)} \\
\cmidrule(lr){3-8} \cmidrule(lr){9-12}
 & & efi & ibb & mfe & oro & quy & vmw & anw & lld & zgh & apd \\
\midrule
\multicolumn{2}{c|}{\textcolor{gray}{\textit{0-shot}}} & \textcolor{gray}{9.1} & \textcolor{gray}{6.3} & \textcolor{gray}{32.2} & \textcolor{gray}{2.3} & \textcolor{gray}{9.9} & \textcolor{gray}{6.4} & \textcolor{gray}{3.2} & \textcolor{gray}{17.5} & \textcolor{gray}{0.8} & \textcolor{gray}{25.5} \\
\midrule
\multirow{4}{*}{1}
  & Random  & 9.4 & 6.6 & 35.5 & 4.1 & 10.8 & 6.8 & 4.9 & 18.1 & 20.0 & \textbf{26.6} \\
  & BM25    & \textbf{10.1} & \textbf{7.5} & \textbf{35.7} & \textbf{7.2} & \textbf{11.2} & \textbf{7.7} & \textbf{5.8} & \textbf{24.1} & \textbf{22.2} & 25.9 \\
  & BM25$-$Random ($\triangle$) & \textcolor{green!60!black}{+0.7} & \textcolor{green!60!black}{+0.9} & \textcolor{green!60!black}{+0.2} & \textcolor{green!60!black}{+3.2} & \textcolor{green!60!black}{+0.4} & \textcolor{green!60!black}{+0.9} & \textcolor{green!60!black}{+1.0} & \textcolor{green!60!black}{+6.0} & \textcolor{green!60!black}{+2.2} & \textcolor{red!70!black}{$-$0.8} \\
  & Bible   & 9.9 & 7.3 & 33.7 & 3.9 & 10.4 & 7.0 & \xmark & \xmark & \xmark & \xmark \\
\midrule
\multirow{4}{*}{5}
  & Random  & 9.6 & 7.9 & 37.3 & 8.8 & 11.7 & 7.3 & 5.3 & 25.6 & 22.1 & 25.2 \\
\bottomrule
\end{tabular}

\caption{eng$\rightarrow$X results with Gemini 2.5 Flash (spBLEU). Formatting conventions as described in \S\ref{app:full_results}.}
\label{tab:bible_vs_indomain_gemini_spbleu}

\end{table*}

\begin{table*}[t]
\centering
\footnotesize

\begin{tabular}{cl|cccccc|cccc}
\toprule
\textbf{Shots} & \textbf{Src} & \multicolumn{6}{c|}{chrF++ (Bible langs)} & \multicolumn{4}{c}{chrF++ (no Bible)} \\
\cmidrule(lr){3-8} \cmidrule(lr){9-12}
 & & efi & ibb & mfe & oro & quy & vmw & anw & lld & zgh & apd \\
\midrule
\multicolumn{2}{c|}{\textcolor{gray}{\textit{0-shot}}} & \textcolor{gray}{23.6} & \textcolor{gray}{22.7} & \textcolor{gray}{54.6} & \textcolor{gray}{14.5} & \textcolor{gray}{29.9} & \textcolor{gray}{21.2} & \textcolor{gray}{21.3} & \textcolor{gray}{31.9} & \textcolor{gray}{2.1} & \textcolor{gray}{44.2} \\
\midrule
\multirow{4}{*}{1}
  & Random  & 23.7 & 23.3 & 55.1 & 19.1 & 30.5 & 21.7 & 21.6 & 32.2 & 16.3 & \textbf{43.9} \\
  & BM25    & \textbf{24.6} & \textbf{24.0} & \textbf{55.4} & \textbf{21.7} & \textbf{30.9} & \textbf{23.5} & \textbf{22.3} & \textbf{33.8} & \textbf{18.1} & \textbf{43.9} \\
  & BM25$-$Random ($\triangle$) & \textcolor{green!60!black}{+0.9} & \textcolor{green!60!black}{+0.7} & \textcolor{green!60!black}{+0.3} & \textcolor{green!60!black}{+2.6} & \textcolor{green!60!black}{+0.4} & \textcolor{green!60!black}{+1.7} & \textcolor{green!60!black}{+0.7} & \textcolor{green!60!black}{+1.6} & \textcolor{green!60!black}{+1.8} & 0 \\
  & Bible   & 24.5 & 23.5 & 53.8 & 17.1 & 30.1 & 22.1 & \xmark & \xmark & \xmark & \xmark \\
\midrule
\multirow{4}{*}{5}
  & Random  & 24.6 & 23.8 & 56.1 & 21.6 & 31.0 & 23.2 & 22.7 & 33.0 & 17.9 & 43.6 \\
\bottomrule
\end{tabular}

\caption{eng$\rightarrow$X results with GPT-4.1 (chrF++). Formatting conventions as described in \S\ref{app:full_results}.}
\label{tab:bible_vs_indomain_gpt4_chrfpp}

\end{table*}

\begin{table*}[t]
\centering
\footnotesize

\begin{tabular}{cl|cccccc|cccc}
\toprule
\textbf{Shots} & \textbf{Src} & \multicolumn{6}{c|}{spBLEU (Bible langs)} & \multicolumn{4}{c}{spBLEU (no Bible)} \\
\cmidrule(lr){3-8} \cmidrule(lr){9-12}
 & & efi & ibb & mfe & oro & quy & vmw & anw & lld & zgh & apd \\
\midrule
\multicolumn{2}{c|}{\textcolor{gray}{\textit{0-shot}}} & \textcolor{gray}{6.9} & \textcolor{gray}{5.8} & \textcolor{gray}{32.9} & \textcolor{gray}{2.7} & \textcolor{gray}{8.0} & \textcolor{gray}{3.8} & \textcolor{gray}{5.9} & \textcolor{gray}{9.8} & \textcolor{gray}{0.3} & \textcolor{gray}{25.2} \\
\midrule
\multirow{4}{*}{1}
  & Random  & 6.0 & 6.6 & 34.5 & 5.1 & 8.1 & 4.2 & 5.5 & 10.3 & 7.2 & \textbf{24.8} \\
  & BM25    & \textbf{6.9} & \textbf{7.1} & \textbf{34.6} & \textbf{8.1} & \textbf{8.5} & \textbf{5.1} & \textbf{5.8} & \textbf{12.3} & \textbf{10.0} & 24.3 \\
  & BM25$-$Random ($\triangle$) & \textcolor{green!60!black}{+0.9} & \textcolor{green!60!black}{+0.5} & \textcolor{green!60!black}{+0.1} & \textcolor{green!60!black}{+3.1} & \textcolor{green!60!black}{+0.4} & \textcolor{green!60!black}{+0.9} & \textcolor{green!60!black}{+0.3} & \textcolor{green!60!black}{+2.0} & \textcolor{green!60!black}{+2.8} & \textcolor{red!70!black}{$-$0.5} \\
  & Bible   & \textbf{6.9} & 6.8 & 32.3 & 3.9 & 7.9 & 4.4 & \xmark & \xmark & \xmark & \xmark \\
\midrule
\multirow{4}{*}{5}
  & Random  & 6.4 & 6.7 & 35.6 & 6.8 & 8.5 & 4.5 & 5.9 & 11.2 & 9.4 & 24.3 \\
\bottomrule
\end{tabular}

\caption{eng$\rightarrow$X results with GPT-4.1 (spBLEU). Formatting conventions as described in \S\ref{app:full_results}.}
\label{tab:bible_vs_indomain_gpt4_spbleu}

\end{table*}

\begin{table*}[t]
\centering
\footnotesize

\begin{tabular}{cl|cccccc|cccc}
\toprule
\textbf{Shots} & \textbf{Src} & \multicolumn{6}{c|}{chrF++ (Bible langs)} & \multicolumn{4}{c}{chrF++ (no Bible)} \\
\cmidrule(lr){3-8} \cmidrule(lr){9-12}
 & & efi & ibb & mfe & oro & quy & vmw & anw & lld & zgh & apd \\
\midrule
\multicolumn{2}{c|}{\textcolor{gray}{\textit{0-shot}}} & \textcolor{gray}{19.9} & \textcolor{gray}{18.9} & \textcolor{gray}{36.5} & \textcolor{gray}{14.4} & \textcolor{gray}{27.0} & \textcolor{gray}{18.7} & \textcolor{gray}{16.8} & \textcolor{gray}{31.8} & \textcolor{gray}{2.0} & \textcolor{gray}{40.7} \\
\midrule
\multirow{4}{*}{1}
  & Random  & 21.8 & 19.8 & 36.6 & 17.3 & 27.1 & 19.0 & 18.2 & 32.3 & 17.6 & 39.7 \\
  & BM25    & \textbf{22.8} & \textbf{21.6} & \textbf{39.3} & \textbf{19.8} & \textbf{28.1} & \textbf{21.2} & \textbf{19.8} & \textbf{33.7} & \textbf{19.6} & \textbf{40.2} \\
  & BM25$-$Random ($\triangle$) & \textcolor{green!60!black}{+1.0} & \textcolor{green!60!black}{+1.8} & \textcolor{green!60!black}{+2.7} & \textcolor{green!60!black}{+2.5} & \textcolor{green!60!black}{+1.1} & \textcolor{green!60!black}{+2.2} & \textcolor{green!60!black}{+1.6} & \textcolor{green!60!black}{+1.4} & \textcolor{green!60!black}{+1.9} & \textcolor{green!60!black}{+0.5} \\
  & Bible   & 22.4 & 20.1 & 37.3 & 16.1 & 26.9 & 19.6 & \xmark & \xmark & \xmark & \xmark \\
\midrule
\multirow{4}{*}{5}
  & Random  & 23.5 & 22.4 & 39.7 & 21.6 & 28.0 & 21.5 & 21.8 & 33.4 & 18.5 & 38.9 \\
\bottomrule
\end{tabular}

\caption{eng$\rightarrow$X results with Llama 3.3 70B (chrF++). Formatting conventions as described in \S\ref{app:full_results}.}
\label{tab:bible_vs_indomain_llama_chrfpp}
\end{table*}

\begin{table*}[t]
\centering
\footnotesize

\begin{tabular}{cl|cccccc|cccc}
\toprule
\textbf{Shots} & \textbf{Src} & \multicolumn{6}{c|}{spBLEU (Bible langs)} & \multicolumn{4}{c}{spBLEU (no Bible)} \\
\cmidrule(lr){3-8} \cmidrule(lr){9-12}
 & & efi & ibb & mfe & oro & quy & vmw & anw & lld & zgh & apd \\
\midrule
\multicolumn{2}{c|}{\textcolor{gray}{\textit{0-shot}}} & \textcolor{gray}{4.0} & \textcolor{gray}{4.1} & \textcolor{gray}{12.9} & \textcolor{gray}{2.4} & \textcolor{gray}{6.0} & \textcolor{gray}{2.5} & \textcolor{gray}{3.5} & \textcolor{gray}{9.6} & \textcolor{gray}{0.3} & \textcolor{gray}{20.2} \\
\midrule
\multirow{4}{*}{1}
  & Random  & 4.4 & 3.8 & 13.0 & 4.4 & 6.1 & 2.5 & 3.5 & 10.2 & 7.2 & 19.3 \\
  & BM25    & \textbf{5.0} & \textbf{5.0} & \textbf{15.7} & \textbf{6.9} & \textbf{6.5} & \textbf{3.5} & \textbf{4.4} & \textbf{12.2} & \textbf{10.2} & \textbf{19.6} \\
  & BM25$-$Random ($\triangle$) & \textcolor{green!60!black}{+0.6} & \textcolor{green!60!black}{+1.2} & \textcolor{green!60!black}{+2.7} & \textcolor{green!60!black}{+2.4} & \textcolor{green!60!black}{+0.4} & \textcolor{green!60!black}{+1.0} & \textcolor{green!60!black}{+0.9} & \textcolor{green!60!black}{+2.0} & \textcolor{green!60!black}{+3.0} & \textcolor{green!60!black}{+0.3} \\
  & Bible   & 4.7 & 4.6 & 13.8 & 3.0 & 5.5 & 2.7 & \xmark & \xmark & \xmark & \xmark \\
\midrule
\multirow{4}{*}{5}
  & Random  & 5.1 & 5.0 & 15.6 & 6.3 & 6.4 & 3.0 & 4.3 & 11.9 & 8.8 & 18.1 \\
\bottomrule
\end{tabular}
\caption{eng$\rightarrow$X results with Llama 3.3 70B (spBLEU). Formatting conventions as described in \S\ref{app:full_results}.}
\label{tab:bible_vs_indomain_llama_spbleu}

\end{table*}

\begin{table*}[t]
\centering
\footnotesize

\begin{tabular}{cl|cccccc|cccc}
\toprule
\textbf{Shots} & \textbf{Src} & \multicolumn{6}{c|}{chrF++ (Bible langs)} & \multicolumn{4}{c}{chrF++ (no Bible)} \\
\cmidrule(lr){3-8} \cmidrule(lr){9-12}
 & & efi & ibb & mfe & oro & quy & vmw & anw & lld & zgh & apd \\
\midrule
\multicolumn{2}{c|}{\textcolor{gray}{\textit{0-shot}}} & \textcolor{gray}{19.8} & \textcolor{gray}{18.5} & \textcolor{gray}{41.2} & \textcolor{gray}{11.8} & \textcolor{gray}{26.1} & \textcolor{gray}{18.7} & \textcolor{gray}{11.7} & \textcolor{gray}{30.5} & \textcolor{gray}{2.9} & \textcolor{gray}{42.6} \\
\midrule
\multirow{4}{*}{1}
  & Random  & 21.2 & 18.9 & 43.5 & 18.7 & 25.4 & 19.2 & 17.7 & 31.3 & 18.0 & 43.5 \\
  & BM25    & \textbf{22.1} & \textbf{20.2} & \textbf{45.7} & \textbf{19.7} & \textbf{26.5} & \textbf{21.4} & \textbf{19.8} & \textbf{33.1} & \textbf{19.5} & \textbf{43.7} \\
  & BM25$-$Random ($\triangle$) & \textcolor{green!60!black}{+0.9} & \textcolor{green!60!black}{+1.3} & \textcolor{green!60!black}{+2.2} & \textcolor{green!60!black}{+1.0} & \textcolor{green!60!black}{+1.1} & \textcolor{green!60!black}{+2.2} & \textcolor{green!60!black}{+2.1} & \textcolor{green!60!black}{+1.8} & \textcolor{green!60!black}{+1.5} & \textcolor{green!60!black}{+0.2} \\
  & Bible   & 21.4 & 18.6 & 43.0 & 15.4 & 25.8 & 19.7 & \xmark & \xmark & \xmark & \xmark \\
\midrule
\multirow{4}{*}{5}
  & Random  & 23.2 & 21.1 & 47.0 & 22.0 & 26.2 & 23.0 & 22.5 & 33.2 & 19.4 & 44.1 \\
\bottomrule
\end{tabular}

\caption{eng$\rightarrow$X results with Gemma 3 27B (chrF++). Formatting conventions as described in \S\ref{app:full_results}.}

\label{tab:bible_vs_indomain_gemma3_27b_chrfpp}
\end{table*}

\begin{table*}[t]
\centering
\footnotesize

\begin{tabular}{cl|cccccc|cccc}
\toprule
\textbf{Shots} & \textbf{Src} & \multicolumn{6}{c|}{spBLEU (Bible langs)} & \multicolumn{4}{c}{spBLEU (no Bible)} \\
\cmidrule(lr){3-8} \cmidrule(lr){9-12}
 & & efi & ibb & mfe & oro & quy & vmw & anw & lld & zgh & apd \\
\midrule
\multicolumn{2}{c|}{\textcolor{gray}{\textit{0-shot}}} & \textcolor{gray}{3.6} & \textcolor{gray}{3.8} & \textcolor{gray}{17.1} & \textcolor{gray}{2.5} & \textcolor{gray}{5.0} & \textcolor{gray}{3.0} & \textcolor{gray}{2.0} & \textcolor{gray}{8.6} & \textcolor{gray}{0.8} & \textcolor{gray}{23.5} \\
\midrule
\multirow{4}{*}{1}
  & Random  & 3.8 & 3.7 & 19.9 & 5.1 & 4.9 & 3.0 & 3.7 & 9.1 & 9.0 & \textbf{24.1} \\
  & BM25    & \textbf{4.8} & \textbf{4.9} & \textbf{22.2} & \textbf{7.4} & \textbf{5.5} & \textbf{4.1} & \textbf{4.8} & \textbf{11.8} & \textbf{11.7} & \textbf{24.1} \\
  & BM25$-$Random ($\triangle$) & \textcolor{green!60!black}{+0.9} & \textcolor{green!60!black}{+1.2} & \textcolor{green!60!black}{+2.4} & \textcolor{green!60!black}{+2.3} & \textcolor{green!60!black}{+0.7} & \textcolor{green!60!black}{+1.0} & \textcolor{green!60!black}{+1.1} & \textcolor{green!60!black}{+2.7} & \textcolor{green!60!black}{+2.8} & 0 \\
  & Bible   & 4.2 & 4.1 & 19.5 & 3.3 & 5.2 & 3.1 & \xmark & \xmark & \xmark & \xmark \\
\midrule
\multirow{4}{*}{5}
  & Random  & 4.9 & 5.1 & 23.1 & 7.3 & 5.5 & 4.4 & 5.8 & 11.4 & 11.3 & 24.4 \\
\bottomrule
\end{tabular}

\caption{eng$\rightarrow$X results with Gemma 3 27B (spBLEU). Formatting conventions as described in \S\ref{app:full_results}.}
\label{tab:bible_vs_indomain_gemma3_27b_spbleu}

\end{table*}
\begin{table*}[t]
\centering
\small

\begin{tabular}{c|cccccccccc}
\toprule
\textbf{Shots} & vmw & zgh & lld & mfe & apd & anw & efi & ibb & oro & quy \\
\midrule
\midrule
\multicolumn{11}{l}{\textit{Gemini 2.5 Flash}} \\
\midrule
0 & \textcolor{gray}{43.5} & \textcolor{gray}{44.0} & \textcolor{gray}{62.2} & \textcolor{gray}{70.8} & \textcolor{gray}{63.6} & \textcolor{gray}{32.3} & \textcolor{gray}{34.5} & \textcolor{gray}{36.5} & \textcolor{gray}{27.8} & \textcolor{gray}{40.6} \\
1 & 44.5 & 44.4 & 63.3 & 71.2 & 64.2 & 34.1 & 35.6 & 38.1 & 28.5 & 41.8 \\
5 & 45.6 & 46.1 & 63.7 & 71.6 & 64.8 & 34.7 & 36.4 & 39.0 & 29.0 & 42.6 \\
10 & 46.3 & 46.3 & 63.9 & 71.6 & 65.0 & 34.9 & 36.5 & 39.1 & 30.6 & 43.0 \\
25 & 46.3 & 47.1 & 64.3 & 71.9 & 65.2 & 35.6 & 36.7 & 40.0 & 31.5 & 43.1 \\
50 & 45.8 & 47.4 & \textbf{64.7} & 71.8 & \textbf{65.7} & 36.4 & 37.2 & 40.4 & 33.4 & 43.6 \\
\bottomrule
\end{tabular}

\caption{X$\rightarrow$eng scaling with random selection (chrF++). Formatting conventions as described in \S\ref{app:full_results}.}
\label{tab:x2eng_stacked_chrfpp}
\end{table*}

\begin{table*}[t]
\centering
\small

\begin{tabular}{c|cccccccccc}
\toprule
\textbf{Shots} & vmw & zgh & lld & mfe & apd & anw & efi & ibb & oro & quy \\
\midrule
\midrule
\multicolumn{11}{l}{\textit{Gemini 2.5 Flash}} \\
\midrule
0 & \textcolor{gray}{23.6} & \textcolor{gray}{24.3} & \textcolor{gray}{41.4} & \textcolor{gray}{54.4} & \textcolor{gray}{42.4} & \textcolor{gray}{14.7} & \textcolor{gray}{17.4} & \textcolor{gray}{17.4} & \textcolor{gray}{10.0} & \textcolor{gray}{20.6} \\
1 & 24.8 & 25.0 & 42.9 & 54.9 & 43.1 & 16.1 & 18.2 & 19.5 & 12.3 & 22.0 \\
5 & 26.4 & 26.8 & 43.6 & 55.5 & 44.8 & 17.7 & 18.9 & 21.4 & 13.7 & 23.5 \\
10 & 27.1 & 27.0 & 44.1 & 55.5 & 44.7 & 17.7 & 19.1 & 21.4 & 14.6 & 23.9 \\
25 & 27.5 & 27.8 & 44.8 & \textbf{56.0} & 45.2 & 18.6 & 19.1 & 22.2 & 15.3 & 24.4 \\
50 & 27.3 & 28.0 & \textbf{45.3} & 55.8 & \textbf{45.9} & 18.9 & 19.7 & 22.6 & 16.4 & 25.0 \\
\bottomrule
\end{tabular}

\caption{X$\rightarrow$eng scaling with random selection (spBLEU). Formatting conventions as described in \S\ref{app:full_results}.}
\label{tab:x2eng_stacked_spbleu}
\end{table*}

\FloatBarrier

\begin{table*}[p]
\centering
\small
\renewcommand{\arraystretch}{0.95}
\begin{tabular}{cl|ccccc|ccccc}
\toprule
\textbf{Shots} & \textbf{Order} & \multicolumn{5}{c|}{spBLEU} & \multicolumn{5}{c}{chrF++} \\
\cmidrule(lr){3-7} \cmidrule(lr){8-12}
 & & vmw & zgh & lld & mfe & apd & vmw & zgh & lld & mfe & apd \\
\midrule
\multicolumn{12}{l}{\textit{Gemini 2.5 Flash}} \\
\midrule
\multicolumn{12}{l}{\textit{BM25 retrieval}} \\
\midrule
\multirow{3}{*}{1} & Default & 7.7 & 22.2 & 24.1 & 35.7 & 25.9 & 28.7 & 28.6 & 43.8 & 57.5 & 44.6 \\
 & Similar first & 7.9 & 22.2 & 23.8 & 35.4 & 25.7 & 28.8 & 28.6 & 43.7 & 57.3 & 44.6 \\
 & Dissimilar first & 7.9 & 21.8 & 24.1 & 35.6 & 25.8 & 28.8 & 28.6 & 43.8 & 57.4 & 44.7 \\
\cmidrule(lr){2-12}
\multirow{3}{*}{5} & Default & 9.7 & 26.1 & 29.0 & 37.4 & 26.3 & 30.9 & 32.0 & 47.6 & 58.6 & 45.7 \\
\bottomrule
\end{tabular}
\caption{Effect of example ordering within BM25 and dense retrieval (eng$\rightarrow$X, FLORES). Left: spBLEU; right: chrF++. For BM25, ``Default'' randomly shuffles the BM25-retrieved top-$k$ examples (the setting used in the main BM25 results, \S\ref{sec:selection}); for dense retrieval, ``Random'' shuffles the retrieved examples. ``Similar first'' places the most-similar example at the start of the prompt (farthest from the query); ``Dissimilar first'' places it immediately before the query. Differences across orderings are negligible ($\leq 0.5$ chrF++).}
\label{tab:ordering_retrieval_stacked}
\end{table*}

\begin{table*}[p]
\centering
\resizebox{\textwidth}{!}{%
\small
\renewcommand{\arraystretch}{0.95}
\begin{tabular}{cl|ccccc|ccccc|ccccc|ccccc}
\toprule
\textbf{Shots} & \textbf{Order} & \multicolumn{10}{c|}{eng$\rightarrow$X} & \multicolumn{10}{c}{X$\rightarrow$eng} \\
\cmidrule(lr){3-12} \cmidrule(lr){13-22}
 & & \multicolumn{5}{c|}{spBLEU} & \multicolumn{5}{c|}{chrF++} & \multicolumn{5}{c|}{spBLEU} & \multicolumn{5}{c}{chrF++} \\
 & & vmw & zgh & lld & mfe & apd & vmw & zgh & lld & mfe & apd & vmw & zgh & lld & mfe & apd & vmw & zgh & lld & mfe & apd \\
\midrule
\multicolumn{22}{l}{\textit{Gemini 2.5 Flash}} \\
\midrule
\multirow{5}{*}{1} & Random & 6.8 & 20.0 & 18.1 & 35.5 & 26.6 & 27.5 & 27.2 & 39.4 & 56.8 & 44.8 & 24.8 & 25.0 & 42.9 & 54.9 & 43.1 & 44.5 & 44.4 & 63.3 & 71.2 & 64.2 \\
 & L2S & 6.9 & 20.2 & 18.3 & 35.3 & 26.5 & 27.5 & 27.4 & 39.6 & 56.7 & 44.7 & 24.9 & 24.8 & 42.9 & 54.6 & 43.2 & 44.6 & 44.3 & 63.4 & 71.0 & 64.2 \\
 & S2L & 6.8 & 19.9 & 18.3 & 35.4 & 26.6 & 27.3 & 26.9 & 39.6 & 56.7 & 44.8 & 24.7 & 25.1 & 43.0 & 54.9 & 43.1 & 44.5 & 44.5 & 63.4 & 71.2 & 64.1 \\
 & Pair-L2S & 6.8 & 20.7 & 18.2 & 35.5 & 26.3 & 27.3 & 27.7 & 39.5 & 56.8 & 44.6 & 25.1 & 25.0 & 42.7 & 54.8 & 43.0 & 44.7 & 44.3 & 63.2 & 71.1 & 64.1 \\
 & Pair-S2L & 6.9 & 20.2 & 18.2 & 35.2 & 26.2 & 27.4 & 27.2 & 39.5 & 56.6 & 44.5 & 25.0 & 25.1 & 42.9 & 54.7 & 43.0 & 44.6 & 44.4 & 63.4 & 71.1 & 64.0 \\
\cmidrule(lr){2-22}
\multirow{5}{*}{5} & Random & 7.3 & 22.1 & 25.6 & 37.3 & 25.2 & 28.2 & 28.7 & 45.1 & 58.3 & 44.3 & 26.4 & 26.8 & 43.6 & 55.5 & 44.8 & 45.6 & 46.1 & 63.7 & 71.6 & 64.8 \\
 & L2S & 7.3 & 21.8 & 25.3 & 37.2 & 25.5 & 28.0 & 28.6 & 44.9 & 58.1 & 44.4 & 26.6 & 26.9 & 43.4 & 55.4 & 44.6 & 45.9 & 46.3 & 63.7 & 71.5 & 64.9 \\
 & S2L & 7.5 & 22.3 & 25.1 & 36.7 & 25.6 & 28.3 & 28.8 & 44.7 & 58.1 & 44.4 & 26.8 & 27.2 & 43.4 & 55.6 & 44.4 & 45.9 & 46.3 & 63.7 & 71.6 & 64.9 \\
 & Pair-L2S & 7.3 & 21.9 & 25.5 & 37.1 & 26.7 & 28.0 & 28.6 & 45.0 & 58.3 & 45.4 & 26.4 & 26.9 & 43.8 & 55.5 & 44.6 & 45.7 & 46.1 & 63.8 & 71.5 & 64.9 \\
 & Pair-S2L & 7.5 & 22.4 & 25.3 & 36.9 & 25.7 & 28.3 & 29.0 & 44.8 & 58.1 & 44.9 & 26.9 & 27.4 & 43.9 & 55.2 & 44.5 & 45.8 & 46.3 & 64.0 & 71.4 & 64.9 \\
\cmidrule(lr){2-22}
\multirow{5}{*}{10} & Random & 8.0 & 22.3 & 26.1 & 37.7 & 26.0 & 28.9 & 29.0 & 45.5 & 58.7 & 45.5 & 27.1 & 27.0 & 44.1 & 55.5 & 44.7 & 46.3 & 46.3 & 63.9 & 71.6 & 65.0 \\
 & L2S & 8.0 & 22.5 & 26.1 & 37.4 & 26.6 & 29.0 & 29.0 & 45.4 & 58.6 & 45.6 & 26.8 & 27.2 & 44.4 & 55.8 & 44.8 & 45.8 & 46.6 & 64.1 & 71.8 & 65.0 \\
 & S2L & 8.3 & 22.6 & 26.0 & 37.8 & 26.5 & 29.1 & 28.9 & 45.4 & 59.0 & 45.8 & 26.4 & 27.5 & 44.1 & 56.1 & 44.7 & 45.9 & 46.5 & 63.8 & 71.9 & 65.1 \\
 & Pair-L2S & 7.7 & 22.3 & 25.7 & 37.3 & 26.9 & 28.6 & 28.9 & 45.1 & 58.5 & 46.1 & 26.8 & 27.5 & 44.5 & 55.9 & 44.4 & 46.0 & 46.7 & 64.1 & 71.8 & 64.8 \\
 & Pair-S2L & 8.1 & 22.5 & 26.2 & 37.7 & 26.8 & 29.2 & 29.3 & 45.5 & 58.7 & 46.0 & 26.0 & 27.2 & 44.3 & 55.6 & 45.1 & 45.5 & 46.5 & 63.9 & 71.7 & 65.3 \\
\cmidrule(lr){2-22}
\multirow{5}{*}{25} & Random & 8.9 & 23.6 & 27.4 & 38.5 & 26.9 & 30.6 & 30.0 & 46.5 & 59.4 & 46.2 & 27.5 & 27.8 & 44.8 & 56.0 & 45.2 & 46.3 & 47.1 & 64.3 & 71.9 & 65.2 \\
 & L2S & 9.1 & 23.7 & 27.2 & 37.9 & 27.8 & 30.8 & 29.9 & 46.3 & 59.1 & 46.9 & 27.6 & 27.5 & 44.5 & 55.9 & 45.5 & 46.8 & 47.0 & 64.1 & 71.8 & 65.6 \\
 & S2L & 9.0 & 23.9 & 27.2 & 38.6 & 27.0 & 30.7 & 30.1 & 46.2 & 59.4 & 46.3 & 27.8 & 27.8 & 45.1 & 55.7 & 45.8 & 46.7 & 47.1 & 64.3 & 71.8 & 65.6 \\
 & Pair-L2S & 9.3 & 23.8 & 27.3 & 37.8 & 27.4 & 30.7 & 29.8 & 46.3 & 59.1 & 46.4 & 27.9 & 27.9 & 44.8 & 56.5 & 45.7 & 46.9 & 47.2 & 64.4 & 72.2 & 65.5 \\
 & Pair-S2L & 8.8 & 23.9 & 27.5 & 38.5 & 27.5 & 30.4 & 30.0 & 46.4 & 59.3 & 46.9 & 27.7 & 27.5 & 44.4 & 56.1 & 45.7 & 46.7 & 46.8 & 64.1 & 71.9 & 65.4 \\
\midrule
\multicolumn{22}{l}{\textit{GPT-4.1}} \\
\midrule
\multirow{5}{*}{1} & Random & 4.2 & 7.2 & 10.3 & 34.5 & 24.8 & 21.7 & 16.3 & 32.2 & 55.1 & 43.9 & 13.2 & 2.8 & 36.7 & 54.7 & 42.5 & 30.2 & 19.4 & 57.2 & 71.1 & 64.2 \\
 & L2S & 4.2 & 7.8 & 10.2 & 34.3 & 24.9 & 21.6 & 16.4 & 32.1 & 55.0 & 44.0 & 13.1 & 2.8 & 36.6 & 54.6 & 42.5 & 30.3 & 19.4 & 57.2 & 71.0 & 64.2 \\
 & S2L & 4.4 & 7.6 & 10.2 & 34.2 & 24.7 & 21.7 & 16.9 & 32.0 & 54.9 & 43.8 & 13.1 & 2.6 & 36.4 & 54.7 & 42.6 & 30.3 & 19.2 & 56.9 & 71.1 & 64.3 \\
 & Pair-L2S & 4.2 & 7.8 & 10.2 & 34.4 & 24.8 & 21.7 & 16.6 & 32.0 & 55.0 & 43.8 & 13.0 & 2.6 & 36.7 & 54.8 & 42.5 & 30.1 & 19.3 & 57.1 & 71.2 & 64.3 \\
 & Pair-S2L & 4.0 & 8.0 & 10.1 & 34.4 & 24.9 & 21.6 & 16.7 & 32.0 & 55.1 & 43.9 & 13.1 & 2.5 & 36.7 & 54.7 & 42.5 & 30.2 & 19.3 & 57.2 & 71.1 & 64.3 \\
\cmidrule(lr){2-22}
\multirow{5}{*}{5} & Random & 4.5 & 9.4 & 11.2 & 35.6 & 24.3 & 23.2 & 17.9 & 33.0 & 56.1 & 43.6 & 14.6 & 3.4 & 38.1 & 55.4 & 44.3 & 31.6 & 20.0 & 58.4 & 71.7 & 65.3 \\
 & L2S & 4.3 & 8.8 & 11.1 & 35.4 & 24.6 & 22.9 & 17.7 & 32.9 & 56.0 & 43.9 & 14.2 & 3.3 & 38.1 & 55.2 & 43.9 & 31.5 & 20.2 & 58.4 & 71.5 & 65.1 \\
 & S2L & 4.4 & 9.1 & 11.0 & 35.5 & 24.3 & 23.1 & 17.7 & 32.8 & 56.0 & 43.6 & 14.6 & 3.2 & 38.5 & 55.4 & 44.1 & 31.6 & 19.9 & 58.7 & 71.6 & 65.1 \\
 & Pair-L2S & 4.5 & 9.1 & 10.9 & 35.6 & 24.6 & 23.1 & 17.7 & 32.9 & 56.2 & 44.0 & 14.6 & 3.4 & 38.1 & 55.4 & 44.1 & 31.7 & 20.2 & 58.4 & 71.6 & 65.2 \\
 & Pair-S2L & 4.5 & 9.3 & 11.0 & 35.7 & 24.5 & 23.2 & 17.7 & 32.9 & 56.2 & 43.6 & 14.5 & 3.3 & 38.2 & 55.3 & 44.1 & 31.6 & 19.9 & 58.5 & 71.6 & 65.1 \\
\cmidrule(lr){2-22}
\multirow{5}{*}{10} & Random & 4.4 & 9.6 & 11.6 & 36.4 & 26.1 & 23.6 & 18.5 & 33.5 & 56.9 & 45.5 & 14.8 & 3.7 & 39.2 & 55.9 & 44.5 & 31.5 & 20.3 & 58.9 & 72.0 & 65.4 \\
 & L2S & 4.2 & 9.5 & 11.6 & 36.0 & 26.1 & 23.3 & 18.4 & 33.7 & 56.8 & 45.5 & 14.5 & 3.7 & 39.1 & 55.7 & 44.8 & 31.6 & 20.4 & 59.0 & 71.9 & 65.5 \\
 & S2L & 4.5 & 9.8 & 11.6 & 35.9 & 25.8 & 23.7 & 18.3 & 33.6 & 56.6 & 45.3 & 14.7 & 3.6 & 39.1 & 55.7 & 44.2 & 31.6 & 20.2 & 59.0 & 71.9 & 65.2 \\
 & Pair-L2S & 4.3 & 9.7 & 11.8 & 36.3 & 25.8 & 23.6 & 18.4 & 33.8 & 56.9 & 45.2 & 14.4 & 3.6 & 39.0 & 55.7 & 44.6 & 31.7 & 20.4 & 58.9 & 71.9 & 65.4 \\
 & Pair-S2L & 4.6 & 9.6 & 11.5 & 36.5 & 25.5 & 23.6 & 18.3 & 33.4 & 56.9 & 44.8 & 14.6 & 3.5 & 38.8 & 56.0 & 44.6 & 31.5 & 20.3 & 58.6 & 72.1 & 65.4 \\
\cmidrule(lr){2-22}
\multirow{5}{*}{25} & Random & 5.1 & 10.0 & 13.3 & 36.6 & 26.1 & 25.3 & 18.8 & 35.3 & 57.2 & 45.7 & 15.4 & 3.5 & 40.4 & 56.6 & 45.6 & 32.2 & 20.3 & 59.8 & 72.4 & 66.0 \\
 & L2S & 4.9 & 9.9 & 13.5 & 36.4 & 26.1 & 25.0 & 18.7 & 35.7 & 57.2 & 46.0 & 15.0 & 3.5 & 39.7 & 56.2 & 45.8 & 32.1 & 20.4 & 59.4 & 72.2 & 66.0 \\
 & S2L & 5.1 & 10.0 & 13.4 & 36.3 & 26.3 & 25.0 & 18.7 & 35.5 & 57.1 & 46.0 & 15.0 & 3.8 & 40.2 & 56.2 & 45.8 & 31.8 & 20.5 & 59.6 & 72.3 & 66.1 \\
 & Pair-L2S & 5.1 & 9.9 & 13.3 & 36.3 & 26.3 & 25.0 & 18.7 & 35.6 & 57.1 & 46.1 & 15.2 & 3.8 & 39.7 & 56.1 & 45.5 & 32.3 & 20.6 & 59.5 & 72.2 & 65.9 \\
 & Pair-S2L & 5.1 & 9.8 & 13.3 & 36.9 & 26.0 & 25.1 & 18.5 & 35.2 & 57.5 & 45.9 & 15.4 & 3.8 & 39.7 & 55.6 & 45.7 & 32.2 & 20.5 & 59.3 & 71.8 & 66.0 \\
\bottomrule
\end{tabular}
}
\caption{Effect of length-based example ordering. Random: no ordering; L2S: Long-to-Short; S2L: Short-to-Long. Pair-L2S/Pair-S2L: ordered by the combined source+target length.}
\label{tab:ordering_length_stacked}
\end{table*}

\end{document}